\documentclass{article}
\usepackage{ijcai21}

\usepackage{times}
\usepackage{subcaption}
\usepackage{soul}
\usepackage{url}
\usepackage[hidelinks]{hyperref}
\usepackage[utf8]{inputenc}
\usepackage{caption}
\usepackage{graphicx}
\usepackage{amsmath,amssymb}
\usepackage{amsthm}
\usepackage{booktabs}
\usepackage{algorithm}
\usepackage{algorithmic}
\usepackage{multirow}
\usepackage{makecell}
\usepackage{array}
\newcolumntype{H}{>{\setbox0=\hbox\bgroup}c<{\egroup}@{}}

\title{Non-I.I.D. Multi-Instance Learning for Predicting Instance and Bag Labels using Variational Auto-Encoder}

\author{
    Weijia Zhang	
    \emails
	weijia.zhang.xh@gmail.com
}

\begin{document}

\maketitle

\begin{abstract}
Multi-instance learning is a type of weakly supervised learning. It deals with tasks where the data is a set of bags and each bag is a set of instances.
Only the bag labels are observed whereas the labels for the instances are unknown. 
An important advantage of multi-instance learning is that by representing objects as a bag of instances, it is able to preserve the inherent dependencies among parts of the objects.
Unfortunately, most existing algorithms assume all instances to be \textit{identically and independently distributed}, which violates real-world scenarios since the instances within a bag are rarely independent.
In this work, we propose the Multi-Instance Variational Auto-Encoder (MIVAE) algorithm which explicitly models the dependencies among the instances for predicting both bag labels and instance labels. 
Experimental results on several multi-instance benchmarks and end-to-end medical imaging datasets demonstrate that MIVAE performs better than state-of-the-art algorithms for both instance label and bag label prediction tasks.
\end{abstract}

\section{Introduction}
Multi-instance learning (MIL) \cite{Dietterich1997} was originally proposed for drug activity prediction where the task is to predict whether a molecule is suitable for binding to a receptor. 
Since each molecule may take many distinct low-energy conformations and scientists only know its suitability for binding at the molecule level, MIL is proposed to model the molecules as bags and the conformations of the molecules as instances, where only the bag labels are provided but the instance labels are unknown to the learner.

Since its inception, MIL has been studied extensively in various applications where the tasks have inherently structured representations or the fine-grained instance labels are expensive to obtain.
For example, MIL has been studied in text categorization \cite{Andrews:2002:SVM:2968618.2968690,Zhou2009,Ji2020} where articles are represented by bags of sentences with only article-level labels, and in image classifications \cite{Chen2006,Ilse2018,Skrede2020,Yao2020} where the images are divided into bags of patches with only image-level labels. 

As many real-world objects are inherently structured, an important advantage of multi-instance learning is that by representing objects as bags of instances it can convey more information than using a flat single-instance representation. Since instances within a bag correspond to parts of an object, they share structural and contextual information inherited from the bag and are unlikely to be independent.
To see this, let us consider the example from drug activity prediction where instances within a bag represent low-energy conformations of the same molecule. 
The instances are evidently not independently and identically distributed (i.i.d.) since all conformations share the same structure of single bonds and only differ in how the single bonds rotate.

Unfortunately, despite the fact that neither the instances within a bag are  independently and identically distributed, nor should they be treated as such \cite{Zhou2009,Zhang2014}, most of the existing MIL algorithms approach the problem by either explicitly assume all instances to be i.i.d. and directly predict the instance labels,
or focus solely on predicting the bag labels by transforming the bags into a single-vector embedded space which buries instance-level information and prohibits instance label prediction.


In this paper, we propose the Multi-Instance Variational Auto-Encoder (MIVAE) algorithm which explicitly models the instances within a bag as non-i.i.d. using a generative model consisted of a shared bag-level latent factor and instance-level latent factors specific to each of the instances (Figure 1). 
By using the bag-level factor to capture the structural and contextual dependencies among the instances and using the instance-level factors to capture the instance-specific variations, MIVAE excels at the tasks of both instance label and bag label prediction.
On the one hand, integrating the shared bag-level factor and the individualized instance-level factors encapsulate sufficient information for predicting the bag labels. 
On the other hand, since the shared structural and contextual dependencies are captured by the bag-level factor and the instance-level factors only capture the variations of the individual instances, the instance-level factors promotes better prediction of the instance labels .

The contributions of this work are three-fold:
(i) We propose the Multi-instance Variational Auto-Encoder (MIVAE) model, to the best of our knowledge, the first neural network-based MIL algorithm that simultaneously models the instances as non-i.i.d.
(ii) We extend the powerful variational auto-encoder framework to MIL and instantiate an implementation of MIVAE for simultaneously inferencing the bag-level and instance-level latent factors and predicting both instance labels and bag labels.
(iii) We empirically demonstrate the effectiveness of MIVAE over classic and neural network-based MIL algorithms using a suite of MIL benchmark datasets and end-to-end medical image diagnosis datasets for both instance label and bag label prediction.


The rest of this paper is organized as follows. We discuss related work in Section 2 and present the proposed MIVAE framework in Section 3. In Section 4 we report the experimental results. Finally, we conclude the paper in Section 5.

\section{Related Work}
Existing MIL algorithms can be roughly divided into two categories depending whether the algorithm operates at the level of instances or the level of bags \cite{Amores2013}.
The first group of algorithms directly tackle the MIL problem at the instance-level by training a discriminative instance classifier which separates positive instances in the positive bags from the negative ones \cite{Andrews:2002:SVM:2968618.2968690,Li2009,Kim2010,Kandemir2014,Haussmann2017}.
The second group of methods operates at the bag-level, either by embedding the bags into a single-vector representation and solve a single-instance learning problem at the embedded space \cite{Chen2006,Wei2017}, or designing multi-instance kernels for measuring similarities between the bags \cite{Gaertner2002,Zhou2009,Xu2019}.


Recently, several neural network-based approaches have been investigated for MIL. 
\cite{Wang2018} extends feed-forward neural networks to MIL by using permutation-invariant pooling operations on the learned instance embeddings. 
\cite{Ilse2018,Shi2020} propose to use the attention mechanism in the pooling layer such that the attention weights can be interpreted as how much the instances contribute to the bag label.
To the best of our knowledge, all of the existing neural network-based MIL algorithms assume the instances to be i.i.d. and the proposed MIVAE is the first deep generative model for MIL.

Several algorithms has approached MIL with generative models which directly predicting the instance labels by either assuming the Gaussian Process \cite{Kim2010,Haussmann2017} or the Dirichlet Process \cite{Kandemir2014}. 
However, all of them explicitly assume the instances within a bag to be i.i.d.
In this work we show that such assumption not only does not agree with the real-world scenarios, but also hurts empirical performance for both bag and instance level label prediction.

Perhaps the work most related to ours is the proposal \cite{Doran2016}, where they model the multi-instance bags as distributions over instances. 
However, their contribution are mainly theoretical while in this work we propose the MIVAE algorithm which is not only applicable to classical MIL tasks with pre-computed features, but also excels at end-to-end learning from weakly-labeled data with neural networks.

\section{Method}
\begin{figure}[t!]
	\centering
		\includegraphics[height=1.5in]{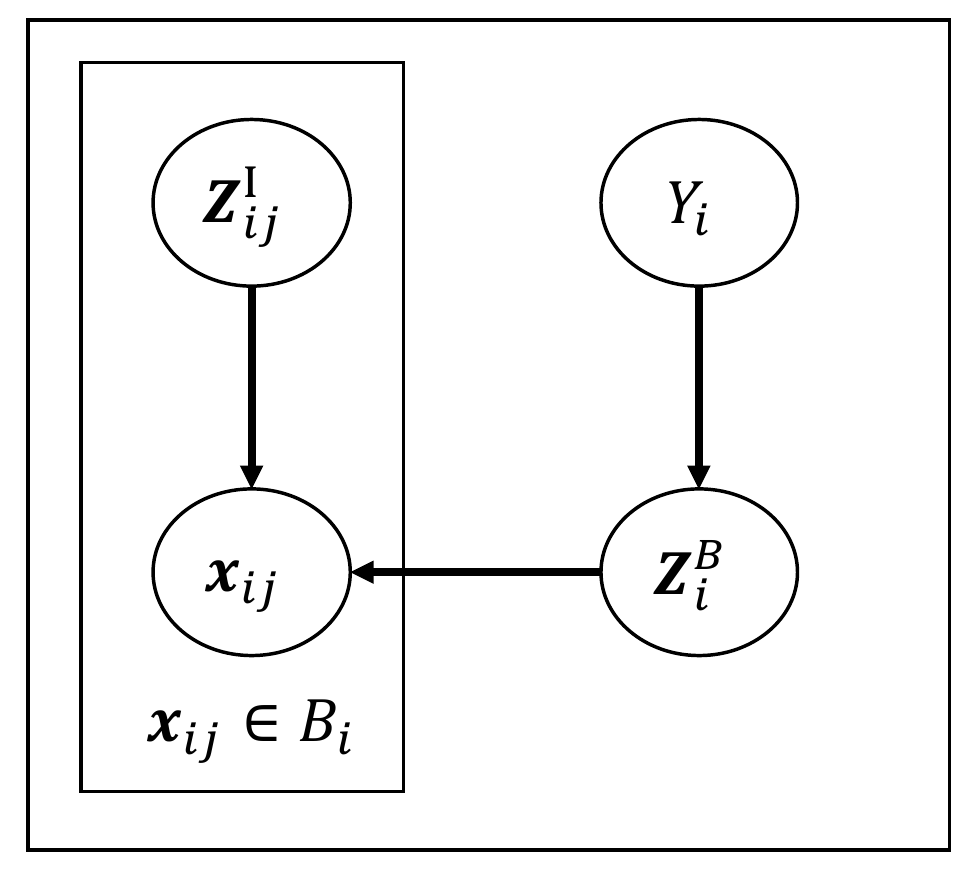}  
	\caption{The generative model for the proposed Multi-Instance Variational Autoencoder (MIVAE). $Y_i$ is the bag label for bag $B_i$, $x_{ij} \in B$ are the instances of $B_i$. $\mathbf{z}^B$ is the bag-level factor shared by all of the instances within $B_i$, and $\mathbf{z}^I_{ij}$ are the instance-level factors specific to each instance.}
	\label{model}
\end{figure}

\begin{figure*}[!t]
	\centering
	\begin{subfigure}[b]{0.95\columnwidth}
		\centering
		\includegraphics[height=1.3in]{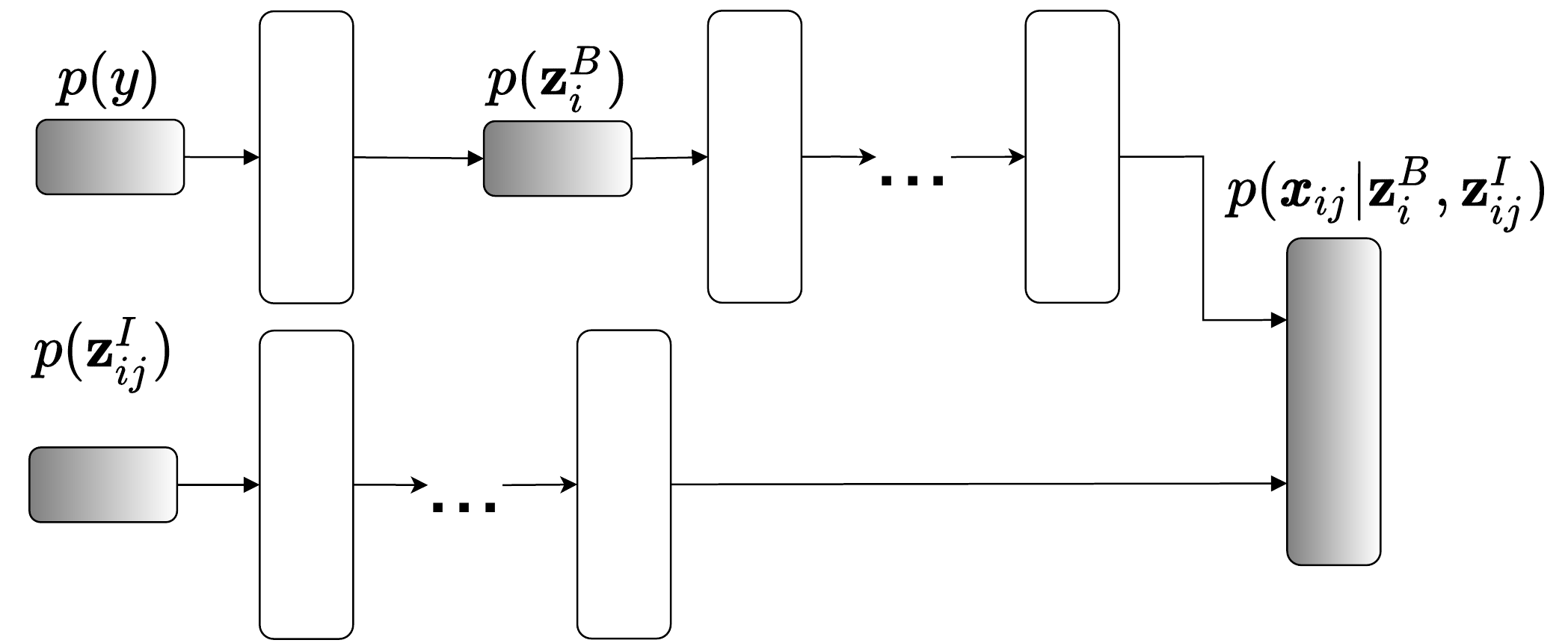}
		\caption{Generative Model.}
	\end{subfigure}
	~
	\begin{subfigure}[b]{0.95\columnwidth}
		\centering
		\includegraphics[height=1.5in]{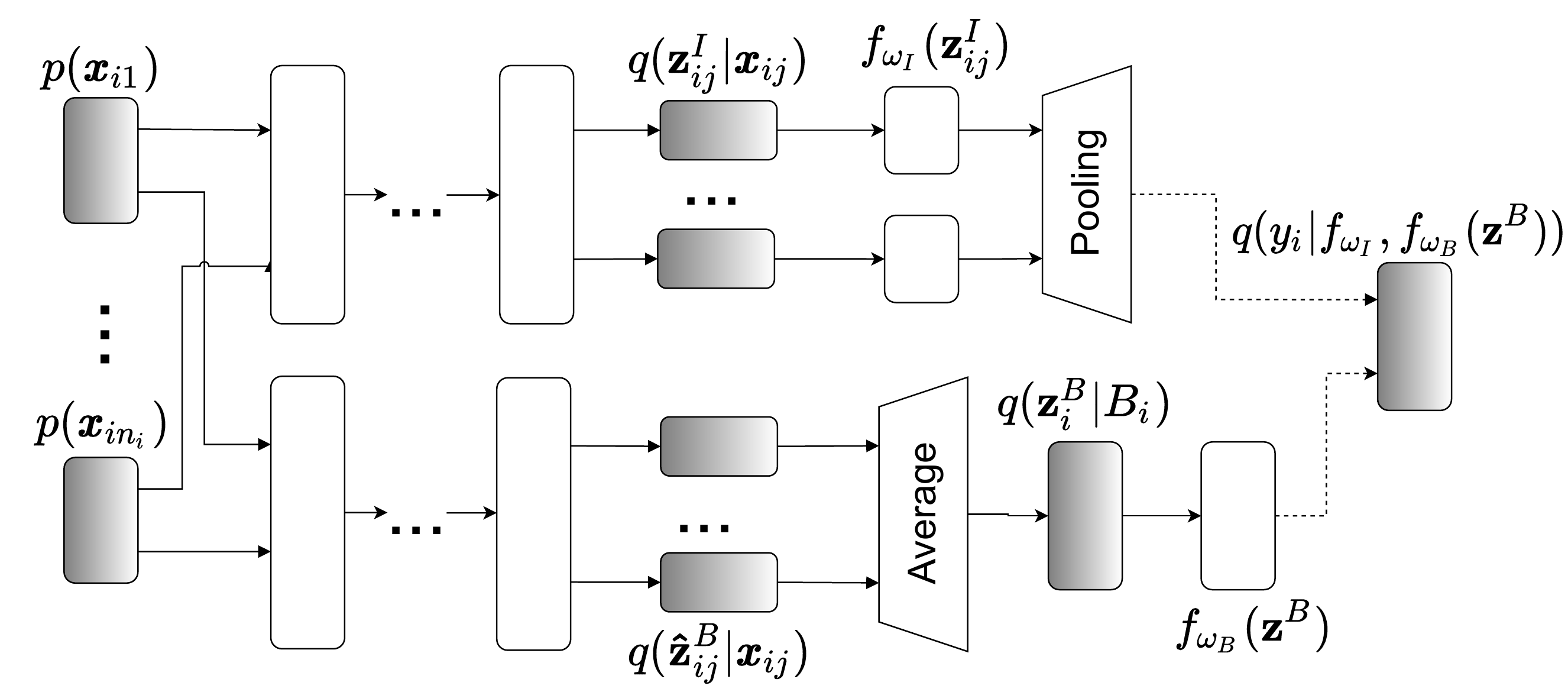}
		\caption{Inference Model.}
	\end{subfigure}
	\caption{Architecture of the model network and the inference network for the Multi-Instance Variational AutoEncoder (MIVAE). White nodes correspond to parametrized deterministic neural network transformations, gray nodes correspond to drawing samples from the respective distribution. Dashed arrows in the inference model represent the auxiliary classifiers $q_{\omega_y} (y|\mathbf{z}^B,\mathbf{z}^I)$.}
	\label{architecture}
\end{figure*}

Let $\mathcal{X} = \mathbb{R}^d$ denote the space of the instances and $\mathcal{Y} = \{0,1\}$ denote the space of the label. 
The learner is given a dataset of $m$ bags multi-instance bags $\mathcal{B} = \{\mathbf{B}_1, \cdots, \mathbf{B}_i,\cdots, \mathbf{B}_m \}$, where each bag $\mathbf{B}_i = \{\pmb{x}_{i1}, \cdots, \pmb{x}_{ij}, \cdots, \pmb{x}_{x_{in_i}} \}$ is a set containing an arbitrary number of  $n_i$ instances with $\pmb{x}_{ij} \in \mathcal{X}$. 
During training, each bag $\mathbf{B}_i$ is provided with a bag label $y_i \in \mathcal{Y}$; however, the label of the instances are unknown. The goal of multi-instance learning as two-fold, to predict both the bag label and the instance labels of unseen bags.
When the context is clear, we drop the subscript $i$ from $\pmb{x}_{ij}$, $\mathbf{B}_i$ and $y_i$ for the conciseness of the notations.

At the heart of this work lies the claim that the instances within the same bag are not independently and identically distributed and the i.i.d. assumption is only applicable at the bag-level.
On the contrary, instances within the same bag share structural and contextual dependencies inherited from the bag and are only independent to each other when the dependencies have been considered.
Apart from drug activity prediction, the dependencies of instances exists widely in MIL applications. 
For example, in text classification instances are paragraphs/sentences of an article and obviously share the styles of the author and contexts of the topic \cite{Ji2020}. 
In medical image diagnosis instances are patches of an organ from a patient which naturally share the pathology features and prognoses of the patient \cite{Skrede2020}.

In order to capture both the shared dependencies and the individual variations of the instances, we propose the MIVAE model as depicted in Figure \ref{model}.
Specifically, it assumes that for each bag the observed instances are generated from two type of latent factors: a bag-level latent factor $\mathbf{z}^B$ shared by all instances, and $n_i$ instance-level latent factors $\mathbf{z}^I_j, j\in\{1,\cdots,n_i\}$ specific to each of the instances.
While the shared bag-level factor $\mathbf{z}^B$ controls the structural and contextual dependencies among the instances, the instance-level factors $\mathbf{z}^I_j$ are responsible for the variations of each instances.
When conditioning on the shared bag-level factor $\mathbf{z}^B$, the instances becomes independent to each other.

At the level of bags, MIL is supervised since each bag is labeled during training. Therefore, we assume the bag-level factor $\mathbf{z}^B$ to be dependent on $y$. 
However, since both the instance labels and how they relate to the bag labels are unknown, assuming fixed generative process would restrict the expressiveness of the model. Therefore, we address the prediction of instance labels at inference time.


\subsection{Multi-Instance Variational Auto-Encoder}

In practice we only observe the bags and bag labels,
our goal is thus to infer the unobservable posterior distributions for both the bag-level and the instance-level latent factors $p_\theta(\mathbf{z}^B|\mathbf{B})$, $p_\theta(\mathbf{z}^I_j|\pmb{x}_j)$, and the posterior distribution of the observable 
$p_\theta(\mathbf{B}|\mathbf{z}^B,\mathbf{z}^I_1,\cdots,\mathbf{z}^I_j)$
where $\theta$ denotes the generative parameters.
Since exact inference of the posterior is intractable due to the fact that both the marginal likelihood and the posterior lack analytical solution, we employ variational autoencoder \cite{Kingma2014} parameterized by neural networks for efficient approximate inference.  

As depicted in Figure \ref{architecture}(a),
we utilize two separately-organized encoding networks $q_{\phi_B}(\mathbf{z}^B|\mathbf{B})$ and $q_{\phi_I}(\mathbf{z}^I_j|\pmb{x}_j)$, where $\phi_B$ and $\phi_I$ denote the parameters, to serve as variational posteriors of the bag-level and instance-level latent factors. 
The encoded latent factors are then feed into a single decoder $p_\theta(\pmb{x}_j|\mathbf{z}^B,\mathbf{z}^I_j)$ for reconstructing the bag of instances. 
Following standard VAE design, the prior distributions of the latent factors $p(\mathbf{z}^B)$ and $p(\mathbf{z}^I_j)$ are chosen as multivariate Gaussians. 
Specifically, the prior distributions of the factors and the generative model are factorized as:
\begin{align*}
	&p(\mathbf{z}^B|y) = \prod_{k=1}^{D_{z_B}} \mathcal{N}(z^{B}_k|f_{y}(y),1);\\
	&p(\mathbf{z}^I_j) = \prod\limits_{k=1}^{D_{z_I}} \mathcal{N}(z^{I}_{jk}|0,1); \nonumber\\
	& p(\pmb{x}_j|\mathbf{z}^B,\mathbf{z}^I_j) = \prod\limits_{k=1}^{d} p(x_{jk}|\mathbf{z}^B,\mathbf{z}^I_j), 
\end{align*}
where $p(x_{jk}|\mathbf{z}^B,\mathbf{z}^I_j)$ is the suitable distribution for the $k$-th feature of the instance, i.e., Gaussians for continuous features or Bernoulli for binary features;
$f_y$ is a function parameterized by neural network using the bag label as input;
$D_{z_B}$ and $D_{z_I}$ are the parameters that determine the dimensionality of the bag-level and instance-level latent factors. 


Comparing to a standard VAE, the inference model of MIVAE needs to overcome two challenges specific to MIL as depicted in Figure \ref{architecture}(b). 
Firstly, apart from inferring the instance-level factors $\mathbf{z}^I$ specific to each instance, MIVAE needs to infer a common bag-level factor $\mathbf{z}^B$ shared by $n_i$ individual instances $\pmb{x}_{j}, j = 1,\cdots,n_i$; however, in standard VAE the latent variables are specific to each of the inputs. 
Secondly, without any instance labels, MIVAE needs to infer the instance labels from the instance-level factors $\mathbf{z}^I_{j}$.


Given instances $\pmb{x}_j$, the instance-level factors $\mathbf{z}^I_j$ can be inferred directly with a standard encoder $q_{\phi_I}(\mathbf{z}^{I}_{j}|\pmb{x}_{j})$ parameterized by neural networks. 
For the shared bag-level factor $\mathbf{z}^B$, a straightforward approach would be directly encoding from the concatenation of all instances within the bag; however, this approach unfortunately violates the permutation-invariant property of the instances, i.e., the encoded bag-level factor may dependent on the ordering of the instances.

To avoid this problem, we design MIVAE to first encode the intermediate bag-level factors $\mathbf{\hat{z}}^{B}_j$ using $q_{\phi_B}(\mathbf{\hat{z}}^{B}_{j}|\pmb{x}_{j})$ for each instance, 
and then utilize a permutation-invariant function for encouraging the intermediate instance-specific factors into a shared bag-level factor. 
Specifically, the variational approximations for the instance-level factors and the intermediate bag-level factors are defined as
\begin{align}
	&q_{\phi_{I}}(\mathbf{z}_j^I|\pmb{x}_j) = \prod_{k=1}^{D_{z_I}} \mathcal{N} (\mu = f_{\phi_{I}}^\psi(\pmb{x}_j), \sigma^2 = f_{\phi_{I}}^\pi(\pmb{x}_j)),\\
	&q_{\phi_{B}}(\mathbf{\hat{z}}^{B}_j|\pmb{x}_j) = \prod_{k=1}^{D_{z_B}} \mathcal{N}(\mu = f_{\phi_{B}}^\psi(\pmb{x}_j), \sigma^2 = f_{\phi_{B}}^\pi(\pmb{x}_j) ),
\end{align}
for $\pmb{x}_j$ with $j=1,\cdots,n_i$, where $f_{\phi_{I}}^\psi, f_{\phi_{B}}^\psi$ and $f_{\phi_{I}}^\pi,  f_{\phi_{B}}^\pi$ are the means and variances for the Gaussian distributions parameterized by neural networks.
Then, the shared bag-level factor is calculated as the mean of the intermediate factors as
\begin{align}
		\resizebox{0.91\linewidth}{!}{
			$
	q_{\phi_B}(\mathbf{z}_B | \mathbf{B}) = \prod\limits_{k=1}^{D_{z_B}} 
	\mathcal{N} ( \frac{1}{n_i}\sum\limits_{j=1}^{n_i} f_{\phi_{B}}^\psi (\pmb{x}_{j}),  \frac{1}{n_i}\sum\limits_{j=1}^{n_i} f_{\phi_{B}}^\pi (\pmb{x}_{j})).
	$
}
\end{align} 
In other words, the encoding neural networks $f_{\phi_{B}}^\psi$ and $f_{\phi_{B}}^\pi$ first infer the means and variances of the posterior distributions for the intermediate factors from the instances $\pmb{x}_{j}$, then the shared bag-level factor is calculated as the average of all the intermediate bag-level factors. 
It is worth noting that the variance of $\mathbf{z}^B$ then become the average of all the intermediate factor variances. 
In this way, as the decoder reconstructs each of the instances using the shared bag-level factor $\mathbf{z}^B$ and individual instance-level factor $\mathbf{z}^I_j$, all of the intermediate bag-level factors are encouraged to converge to the same value during optimization. 
Therefore, $\mathbf{z}^B$ will become a shared bag-level latent factor common to all $\pmb{x}_{j}$, while each $\mathbf{z}^I_j$ captures as the instance-specific variations.  

The generative and inference models of MIVAE can be learned by maximizing the marginal likelihood over the bags
\begin{align}
	\log p_\theta(\mathbf{B},y) &= \mathbb{E}_{q_{\phi_B} q_{\phi_I}} [\log \frac{p_\theta(\mathbf{B},y, \mathbf{z}^B, \mathbf{z}^I)} {q_\phi(\mathbf{z}^B, \mathbf{z}^I |\mathbf{B},y)}] \nonumber\\
	& + \mathbb{E}_{q_{\phi_B} q_{\phi_I}} [\log \frac{q_\phi(\mathbf{z}^B, \mathbf{z}^I |\mathbf{B},y)} {p_\theta(\mathbf{z}^B, \mathbf{z}^I |\mathbf{B},y)}] ,
	\label{likelihood}
\end{align}
where the second term is the Kullback-Leibler (KL) divergence from the variational approximation to the true posterior. 
Because the KL-divergence is non-negative, the first term serves as a lower bound of the marginal likelihood which is also called the evidence lower bound (ELBO).
Maximizing the marginal likelihood is equivalent to maximizing the ELBO, which can be re-written as
\begin{align}
	\mathcal{L}_{\textrm{ELBO}}(\mathbf{B},y) = & \mathbb{E}_{q_{\phi_B}{q_{\phi_I}}} [\log p_\theta (\mathbf{B},y|\mathbf{z}^I, \mathbf{z}^B)] \nonumber\\
	& -  D_{KL} [q_{\phi_B}(\mathbf{z}^B|\mathbf{B})|| p_{\theta_B}(\mathbf{z}^B|y)] \nonumber\\
	& -  \sum\limits_{j=1}^{n_i} D_{KL} [q_{\phi_I}(\mathbf{z}^I_j|\pmb{x}_j)|| p_{\theta_I}(\mathbf{z}^I_j)].
\end{align}

A missing piece of the puzzle is how to predict the instance labels from the inferred instance-level factors $\mathbf{z}^I$ and predict the bag labels from both bag-level and instance-level factors. 
Instead of assuming fixed generative distributions which may restrict the applicability of the model, we employ an auxiliary classifier $q_{\omega_I}(y|\mathbf{z}^B, \mathbf{z}^I_1, \cdots,  \mathbf{z}^I_{n_i} )$ attached to the evidence lower bound for end-to-end prediction of both instance labels and bag labels. 
Using the auxiliary classifier allows MIVAE to have the flexibility of utilizing any permutation-invariant multi-instance pooling techniques, i.e., max-pooling, LSE-pooling, and attention pooling \cite{Ilse2018}. 
In order to demonstrate that disentangling the dependencies among instances into $\mathbf{z}^B$ and using only the instance-level factors  $\mathbf{z}^I_j$ promotes better instance label prediction performance, we use the straightforward max-pooling operation which is defined as
\begin{align}
	f_{I} = \max\limits_{i=1,\cdots,n_i}\{f_{\omega_I}(\mathbf{z}^I_1),\cdots,f_{I}(\mathbf{z}^I_{n_i})\},
\end{align}
where $f_{\omega_I}$ is a neural network with a sigmoid activation to map the instance-level factors to probabilities. 
Then, the auxiliary for predicting the bag label can be expressed as
\begin{align}
q_{\omega} (y|\mathbf{z}^B,\mathbf{z}^I_1,\cdots, \mathbf{z}^I_{n_i}) = q_{\omega} (y|f_{\omega_B}(\mathbf{z}^B), f_{I}),
\end{align}
where $f_{\omega_B}$ is also a parameterized neural network that outputs probabilities.
Finally, the MIVAE objective can be expressed as the sum of the ELBO and the auxiliary classifier,
\begin{align}
	\resizebox{0.91\hsize}{!}{%
	$\mathcal{L}_{\text{MIVAE}} =  \mathcal{L}_{\textrm{ELBO}}(\mathbf{B},y)
	 + \alpha \mathbb{E}_{q_{\phi_{B}} q_{\phi_{I}}} [\log q_{\omega}(y|\mathbf{z}^B,\mathbf{z}^I_1,\cdots, \mathbf{z}^I_{n_i})],
	\label{loss_function}
	$%
}
\end{align}
where $\alpha$ is a weighting parameter of the auxiliary objective. 
Using the re-parameterization trick \cite{Kingma2014}, the objective of MIVAE can be efficiently optimized using stochastic gradient descent. 

For predicting the instance labels of unseen bags, we use the trained instance-level encoder $q_{\phi_I}(\mathbf{z}^I_j|\pmb{x}_j)$ to infer the mean of the posteriors for each $\pmb{x}_j \in B$, and use the intermediate output $f_{\omega_I}(\mathbf{z}^I_{nj})$ from the auxiliary classifier. 
For predicting the bag labels, we use the bag-level encoder $q_{\phi_B}(\mathbf{z}^B|\mathbf{B})$ to infer the mean of the bag-level latent factor $\mathbf{z}^B$, and directly use the output of the auxiliary classifier $q_{\omega} (y|\mathbf{z}^B,\mathbf{z}^I_1,\cdots, \mathbf{z}^I_{n_i})$ for prediction.

\section{Experiments}
We empirically evaluate MIVAE against state-of-the-art MIL algorithms for both bag label prediction and instance label prediction tasks.
Firstly, we compare against SVM-based algorithms including mi-SVM \cite{Andrews:2002:SVM:2968618.2968690}, miFV \cite{Wei2017}, mi-Graph \cite{Zhou2009}, and KI-SVM \cite{Li2009}. 
Secondly, we compare against neural network-based algorithms including mi-Net and MI-Net \cite{Wang2018}, and AttentionMIL \cite{Ilse2018}. 
Thirdly, we compare against generative MIL algorithms for predicting instance labels, including Gaussian Process MIL (GPMIL) \cite{Kim2010}, Dirichlet Process MIL (DPMIL) \cite{Kandemir2014}, and Variational Gaussian Process MIL (VGPMIL) \cite{Haussmann2017}.

We implement MIVAE using PyTorch and provide the source code, datasets, network structures, and parameter grids in the supplementary. Unless otherwise stated, the reported results are obtained from averaging the test results of 10 times repeated 10-fold cross validation as in previous MIL literature. Training for MIVAE are conducted for 100 epochs and parameters are tuned using validation loss evaluate with 10\% of the training bags. 
All experiments are conducted using a PC with a AMD Ryzen 3700x CPU with 32GB RAM and a single Nvidia GTX 1080Ti GPU with 11GB memory.

\subsection{Bag Classification: MIL Benchmarks}
\begin{table}[!t]
	\tiny
	\centering
	\caption{Bag label prediction results on five benchmark MIL benchmarks. The highest average accuracy is marked in bold.}	
	\begin{tabular}{ l | c c c c c} 
		\hline
		Method & Musk1 & Musk2 & Fox & Tiger & Elephant \\
		\hline
		mi-SVM & .874$\pm$.120 & .836$\pm$.088 & .582$\pm$.102 & .789$\pm$.089 & .820$\pm$.073 \\
		MILES & .842$\pm$.081 & .838$\pm$.095 & \textbf{.760$\pm$.045} & .840$\pm$.081 & \textbf{.891$\pm$.053}\\
		mi-Graph &  .889$\pm$.073 & \textbf{.903$\pm$.086} & .616$\pm$.079 & .801$\pm$.083 & .869$\pm$.078 \\
		miFV & .878$\pm$.013 & .868	$\pm$.094 & .621$\pm$.109 & .813$\pm$.083 & .852$\pm$.081 \\
		\hline
		mi-Net & .892$\pm$.040 & .858$\pm$.048 & .615$\pm$.043 & .839$\pm$.064& .868$\pm$.052\\		
		AttentionMIL & .900$\pm$.050 & .863$\pm$.042 & .603$\pm$.059 & .845$\pm$.048 & .857$\pm$.057\\
		\hline
		$\text{MIVAE}$ & \textbf{.904$\pm$.050} & .890$\pm$0.62 & .626$\pm$.055 & \textbf{.850$\pm$.051} & .870$\pm$.064\\
		\hline
	\end{tabular}
	\label{benchmark}
\end{table}

\begin{table*}[!hbt]
	\centering
	\caption{Average test Area Under the Precision-Recall Curve (AUC-PR) of the 20 Newsgroup datasets. The results for the compared methods are obtained from the literature where standard deviations are not reported. The highest average AUC-PR is marked in bold. The last row summarizes the average of the 20 datasets.}
	\begin{tabular}{ l| H c c c c c  H | H c H }
		\hline
		& MI-SVM & mi-SVM & KI-SVM  & GPMIL  & DPMIL  & VGPMIL & AttenMIL & MIVAE-att & MIVAE & MIVAE-max-emb \\
		\hline
		alt.atheism & 0.38 & 0.53 & 0.68 & 0.44 & 0.67 & 0.70 &  & 0.72 & \textbf{0.745$\pm$.030} & \textbf{0.745$\pm$.029} \\
		comp.graphics & 0.07& 0.65 & 0.47 & 0.49 & 0.79 & 0.79 &  0& 0.828$\pm$0.038 & \textbf{0.800$\pm$.042} &79.0$\pm$3.6\\
		comp.os.ms-windows.misc & 0.03 & 0.42 & 0.38 & 0.36 & 0.51 & 0.52 & & 0.487$\pm$0.033 & \textbf{0.548$\pm$.038} & 53.7$\pm$4.0 \\
		comp.sys.ibm.pc.hardware & 0.10 & 0.57 & 0.31 & 0.35 & 0.67 & 0.70 & & 0.668$\pm$0.039 & \textbf{0.711$\pm$.034}\\
		comp.sys.mac.hardware & 0.27 & 0.56 & 0.39 & 0.54 & 0.76 & \textbf{0.79} & & &0.783$\pm$.035 \\
		comp.windows.x & 0.04 & 0.56 & 0.37 & 0.36 & 0.73 & 0.69 & & & \textbf{0.754$\pm$.032}\\
		misc.forsale & 0.10 & 0.31 & 0.29 & 0.33 & 0.45 & 0.54 & & &\textbf{0.553$\pm$.334} \\
		rec.autos & 0.34 & 0.51 &  0.45 & 0.38 & 0.76 & 0.71 & & & \textbf{0.720$\pm$.024}\\
		rec.motorcycles & 0.27 & 0.09 & 0.52 & 0.46 & 0.69 & 0.76 & & &\textbf{0.766$\pm$.029} \\
		rec.sport.baseball & 0.22 & 0.18 & 0.52 & 0.38 & 0.74 & 0.76 & & & \textbf{0.764$\pm$.036}	\\
		rec.sport.hockey & 0.75 & 0.27 & 0.66 & 0.43 & 0.91 & \textbf{0.94} & & & 0.925$\pm$.020\\
		sci.crypt & 0.32 & 0.57 & 0.47 & 0.31 & 0.68 & \textbf{0.82} & & & 0.773$\pm$.036\\
		sci.electronics & 0.34 & 0.83 & 0.42 & 0.71 & 0.90 & 0.92 & & & \textbf{0.928$\pm$.020} \\
		sci.med & 0.44 & 0.37 & 0.55 & 0.32 & 0.73 & 0.73 & & & \textbf{0.745$\pm$.025}\\
		sci.space & 0.20 & 0.46 & 0.51 & 0.32 & 0.70 & 0.74 &  &0.738$\pm$0.031 & \textbf{0.748$\pm$.027}\\
		soc.religion.christian & 0.40 & 0.05 & 0.53 & 0.45 & 0.72 & 0.73 & & & \textbf{0.753$\pm$.035}\\
		talk.politics.guns & 0.01 & 0.57 & 0.43 & 0.38 & 0.64 & \textbf{0.72} & & & 0.714$\pm$.038 \\
		talk.politics.mideast & 0.60 & 0.77 & 0.60 & 0.46 & 0.80 & \textbf{0.87} & & 0.752$\pm$0.05\ & 0.840$\pm$.020\\
		talk.politics.misc  & 0.30 & 0.61 & 0.50 & 0.29 & 0.60 & 0.64 & & 0.665$\pm$0.037 & \textbf{0.650$\pm$.044}\\
		talk.religion.misc & 0.04 & 0.08 & 0.32 & 0.32 & 0.51 & 0.49 & & & \textbf{0.525$\pm$.035} \\
		\hline
		W/T/L & 20/0/0 & 20/0/0 & 20/0/0 & 20/0/0 & 20/0/0 & 15/0/5 & -\\
		\hline
	\end{tabular}
	\label{text}
\end{table*}
We first study the performance of MIVAE for the bag label predictions on five MIL benchmark datasets including the drug activity prediction datasets Musk1 and Musk2, and three content-based image retrieval tasks Fox, Tiger, and Elephant. Following previous works, we report the average accuracy and the standard deviation obtained from repeating 10-fold cross validation for 10 times.

Since these benchmark datasets only contains a small number of bags (approximately 100 bags for the Musk datasets and 200 bags for the image retrieval datasets) and the pre-computed feature vectors, we do not expect MIVAE to significantly outperform existing methods since its deep generative model may not be well suited for these type of tasks. 
However, from Table \ref{benchmark} we can see that the performance of MIVAE is competitive to the compared baselines. Specifically, MIVAE achieves the highest average accuracy on Musk1 and Tiger datasets, and performs competitively on the others.
The results demonstrate that the model of MIVAE is sucessful for handling the uncertainties of multi-instance learning.

\subsection{Instance Classification: 20 NewsGroups}

To study the performance for predicting instance labels, we evaluate MIVAE against baselines using the 20 Newsgroup dataset \cite{Zhou2009}. 
It contains a collection of 20 datasets where each of them is consisted of 100 bags.
Each bag contains approximately 40 instances of articles from 20 different topics where each instance represents one article described by the top 200 TF-IDF features. A bag is considered to be positive if at least one of its instances belongs to a specific topic.
The labels of the bags are distributed evenly; however, in each positive bag only approximately 3\% of the instances are positive. Because of the proportions of positive instances are small, the dataset has been widely used as a standard benchmark for instance label prediction performances.

Following previous work, we report results obtained from repeating 10-fold cross validation for 10 times using the train/test splits provided in \cite{Zhou2009}.  
Because the ground truth instance labels are highly imbalanced, we use Area Under the Precision-Recall Curve (AUC-PR) instead of the area under the receiver operating characteristic curve. Furthermore, in order to demonstrate that the advantage of MIVAE lies in its model instead of parameter tuning, the parameters of MIVAE are selected using only the first of the 20 Newsgroup datasets (\textit{alt.atheism}), and the same parameters are applied across all the rest of the  datasets.

As shown by the Table \ref{text}, without tuning parameters for each specific dataset, MIVAE outperforms the compared baselines on 15 out of 20 datasets for predicting instance labels despite the small number of training bags.
Since all compared algorithms assume the instances to be i.i.d., 
these results confirm the advantage of MIVAE for modeling the dependencies among the instances.

\begin{table}[!t]
	\centering
	\caption{Bag label and instance label prediction results on the test data of the Colon Cancer dataset. Experiments were repeated for 5 times and the average accuracy $\pm$ standard deviation of 10-fold cross validations are reported.}
	\begin{tabular}{l| c H H H c}
		\hline
		Method & Accuracy & Precision & Recall & F-Score & AUC-PR \\
		\hline
		mi-Net & 0.824$\pm$0.021 & 0.866$\pm$0.017 & 0.816$\pm$0.031 & 0.813$\pm$0.023 & 0.491$\pm$0.028\\
		MI-Net & 0.845$\pm$0.015 & 0.884$\pm$0.014 & 0.753$\pm$0.020 & 0.839$\pm$0.017 & 0.466$\pm$0.031\\
		\hline
		AttentionMIL & 0.900$\pm$0.021 & 0.953$\pm$0.014 &0.855$\pm$0.017 & 0.901$\pm$0.011 & 0.500$\pm$0.034\\
		Gated-AttentionMIL & 0.895$\pm$0.026 & 0.944$\pm$0.016& 0.851$\pm$0.035 & 0.893$\pm$0.022 & 0.513$\pm$0.023\\
		\hline
		MIVAE&  \textbf{0.925$\pm$0.014} & & & & \textbf{0.747$\pm$0.032}\\
		\hline
	\end{tabular}
	\label{histopathology}
\end{table}
\begin{figure*}[!t]
	\centering
	\begin{subfigure}{0.18\textwidth}
	\includegraphics[height=1.2in]{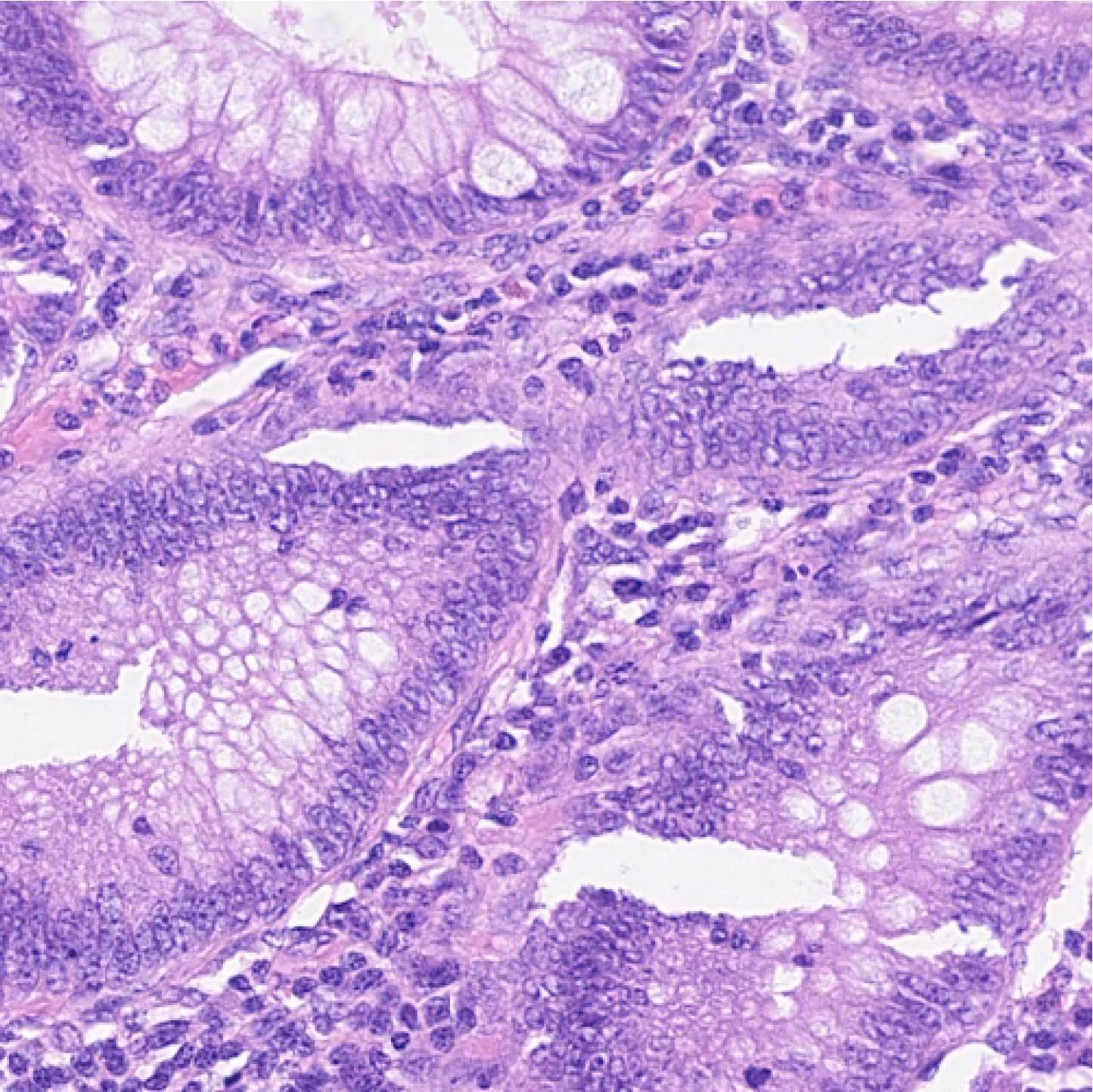}
	\end{subfigure}
	\begin{subfigure}{0.18\textwidth}
	\includegraphics[height=1.2in]{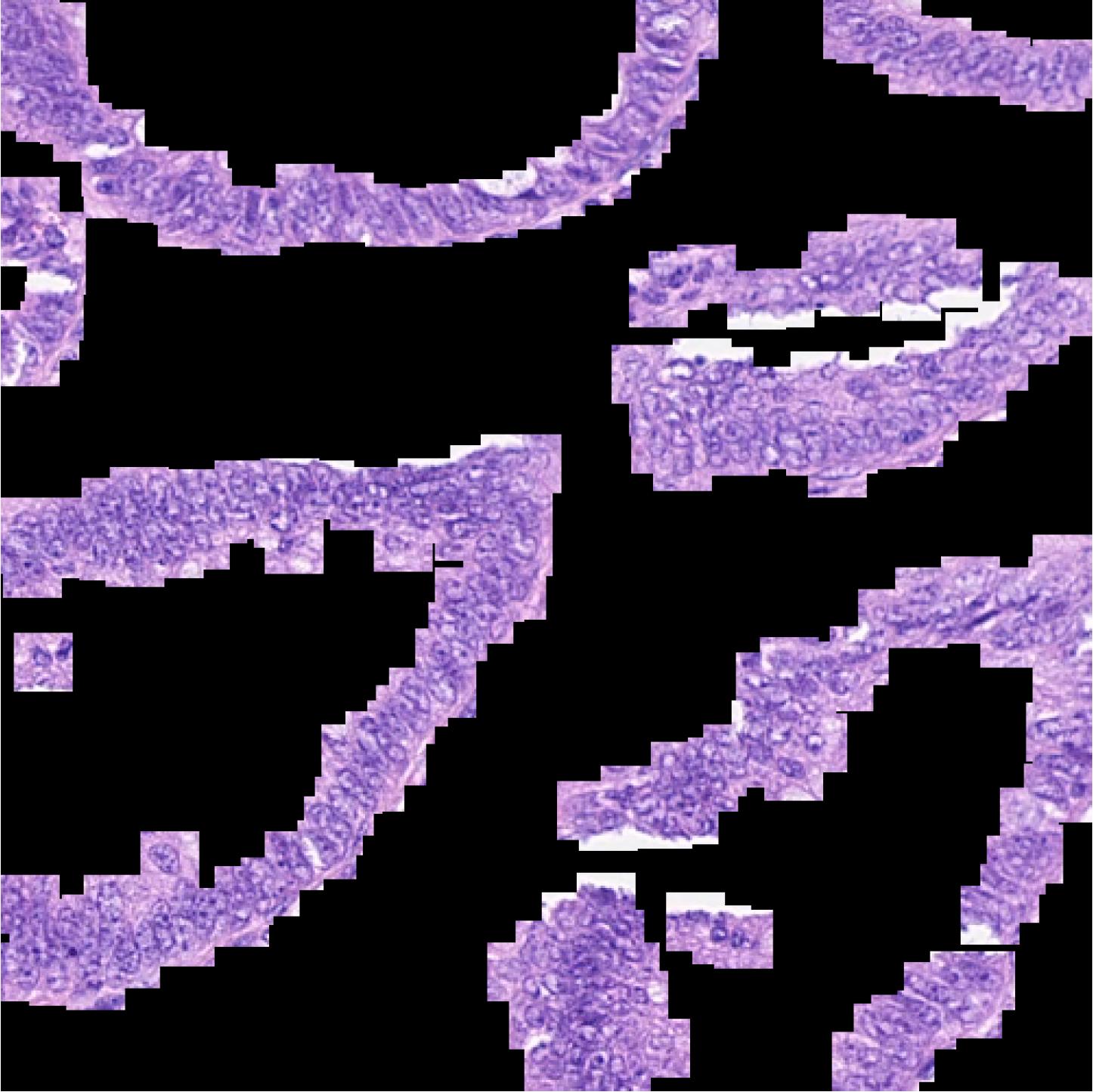}
	\end{subfigure}
	\begin{subfigure}{0.18\textwidth}
		\includegraphics[height=1.2in]{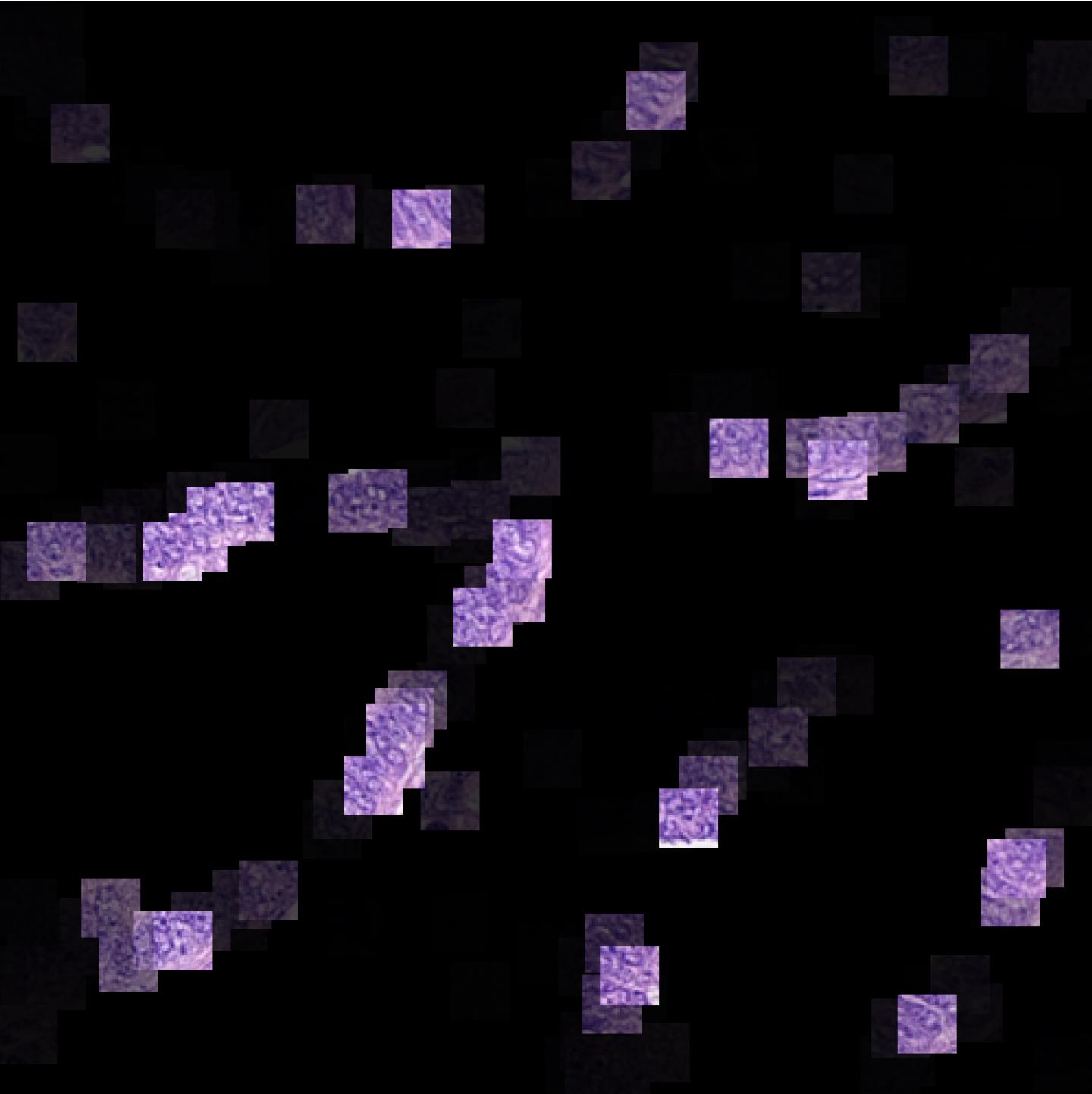}
	\end{subfigure}
	\begin{subfigure}{0.18\textwidth}
		\includegraphics[height=1.2in]{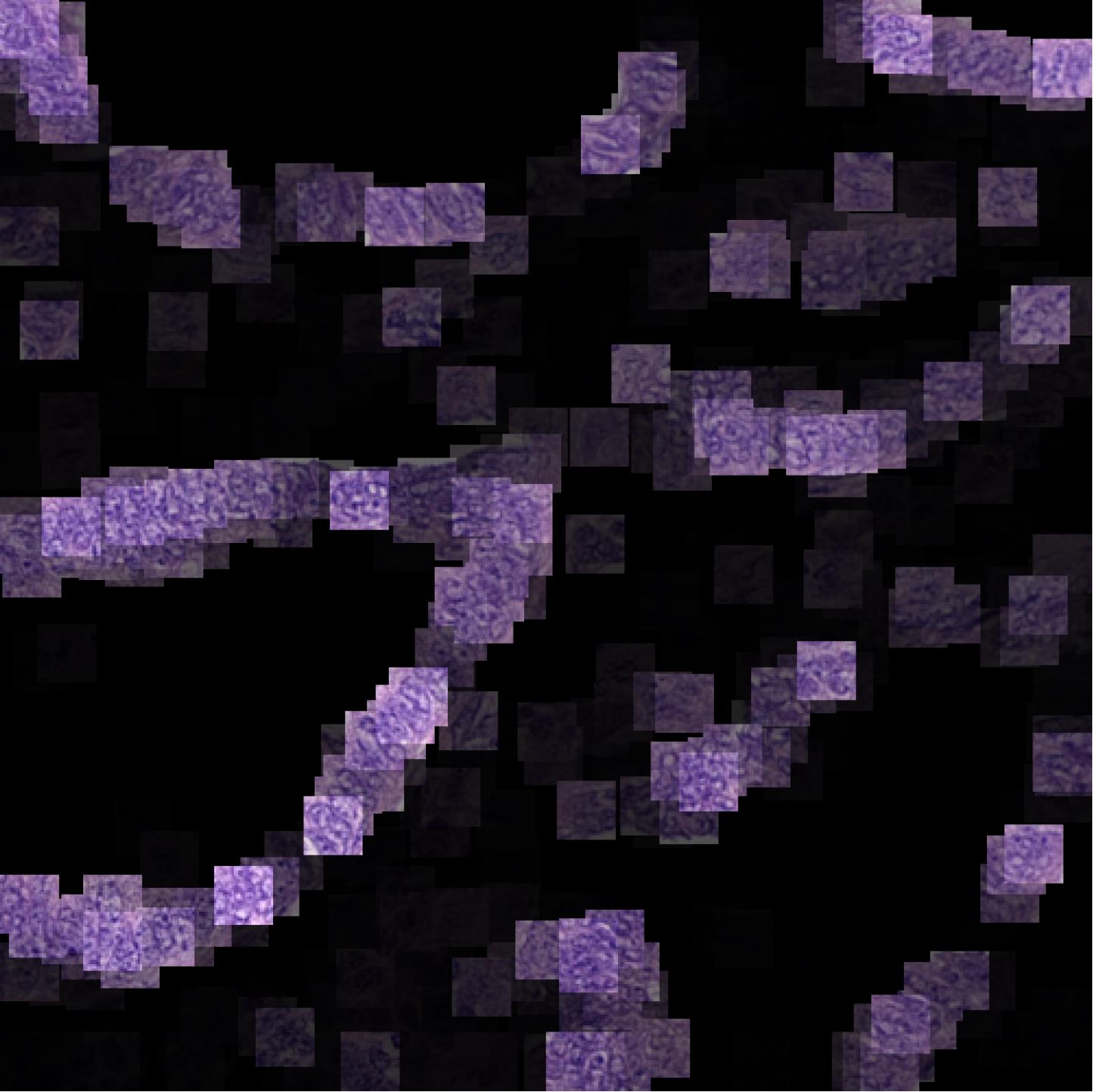}
	\end{subfigure}
	\begin{subfigure}{0.18\textwidth}
		\includegraphics[height=1.2in]{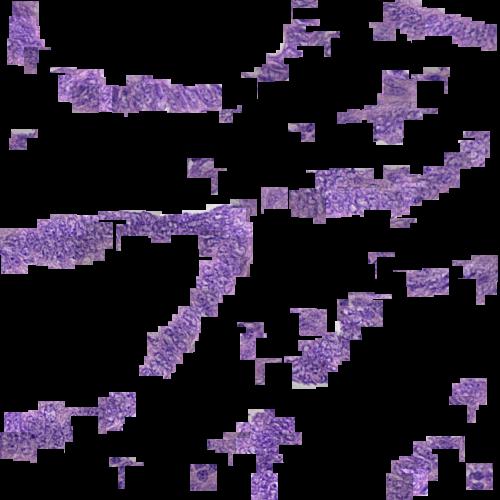}
	\end{subfigure}
	\\
	\vspace{1mm}
	\begin{subfigure}{0.18\textwidth}
	\includegraphics[height=1.2in]{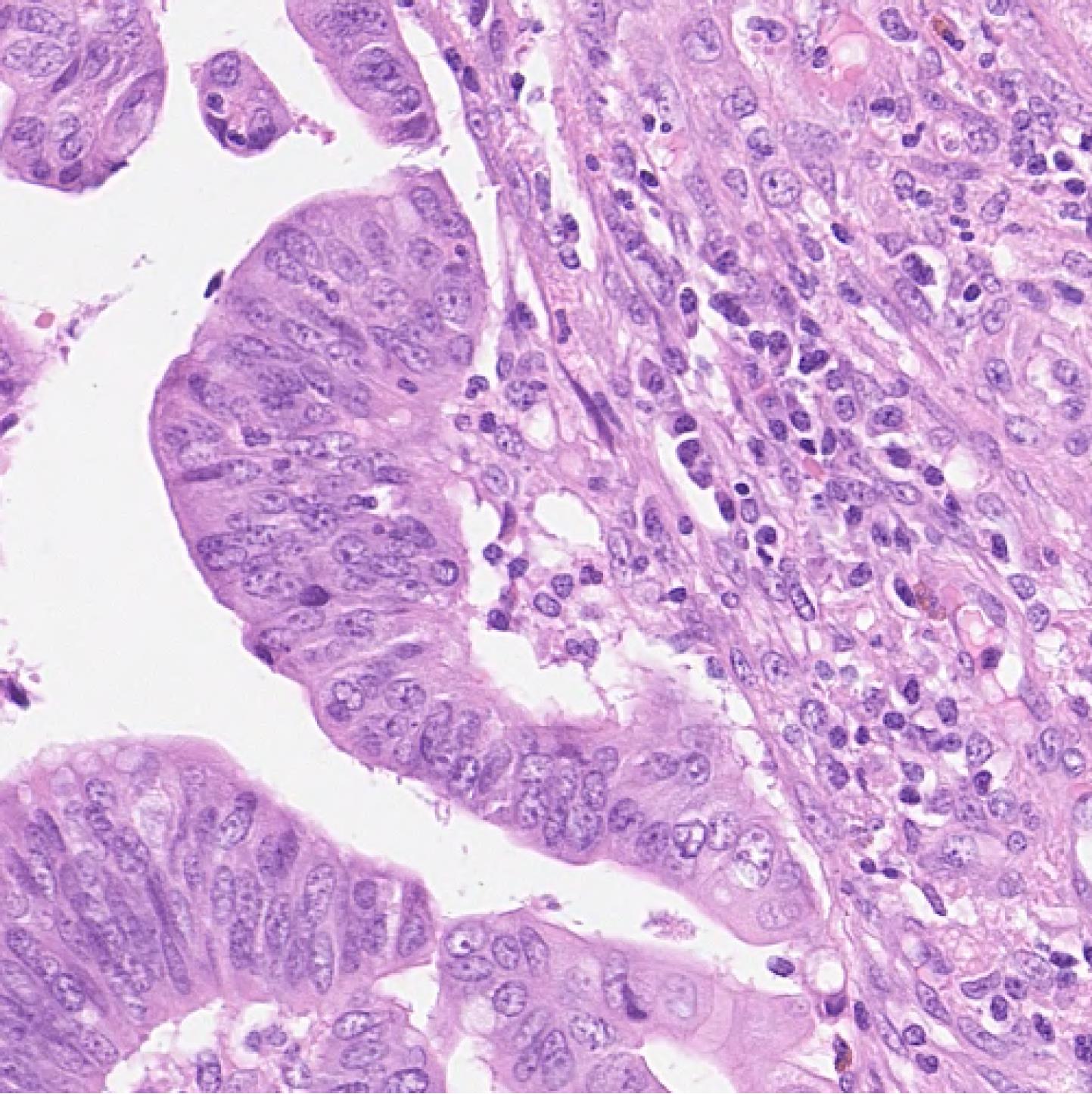}
	\caption{Original image}
	\end{subfigure}
	\begin{subfigure}{0.18\textwidth}
		\includegraphics[height=1.2in]{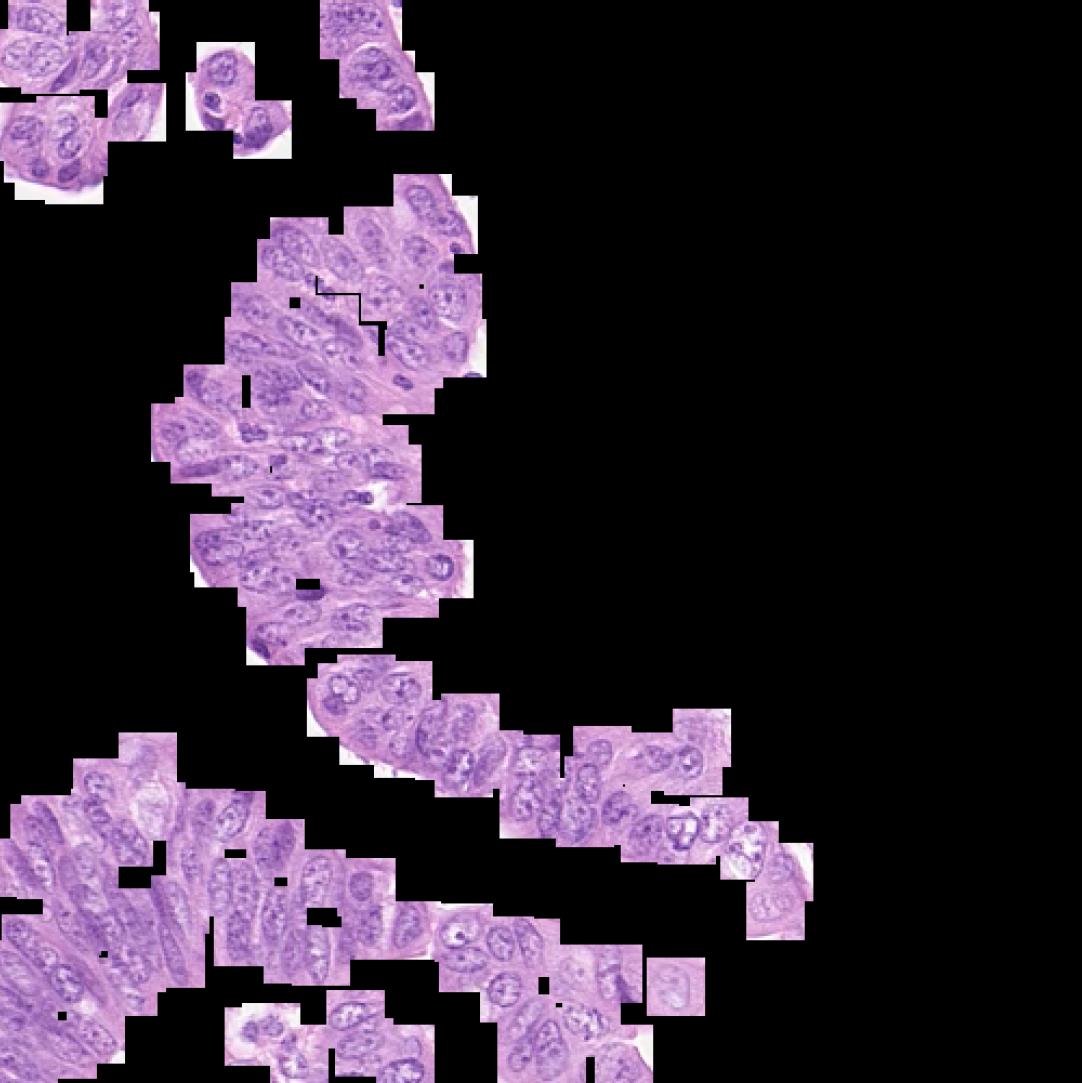}
		\caption{Ground truth}
	\end{subfigure}
	\begin{subfigure}{0.18\textwidth}
		\includegraphics[height=1.2in]{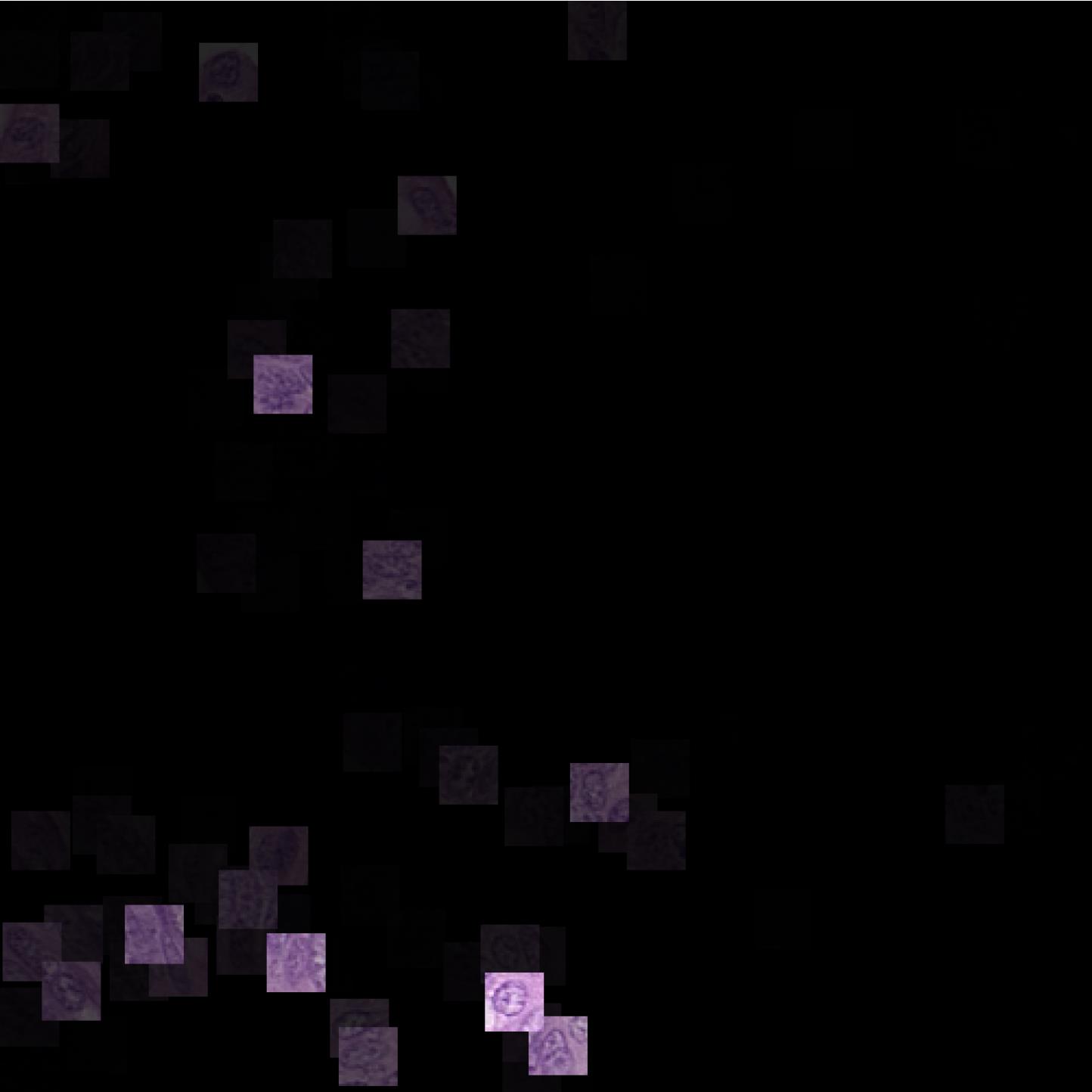}
		\caption{mi-Net}
	\end{subfigure}
	\begin{subfigure}{0.18\textwidth}
		\includegraphics[height=1.2in]{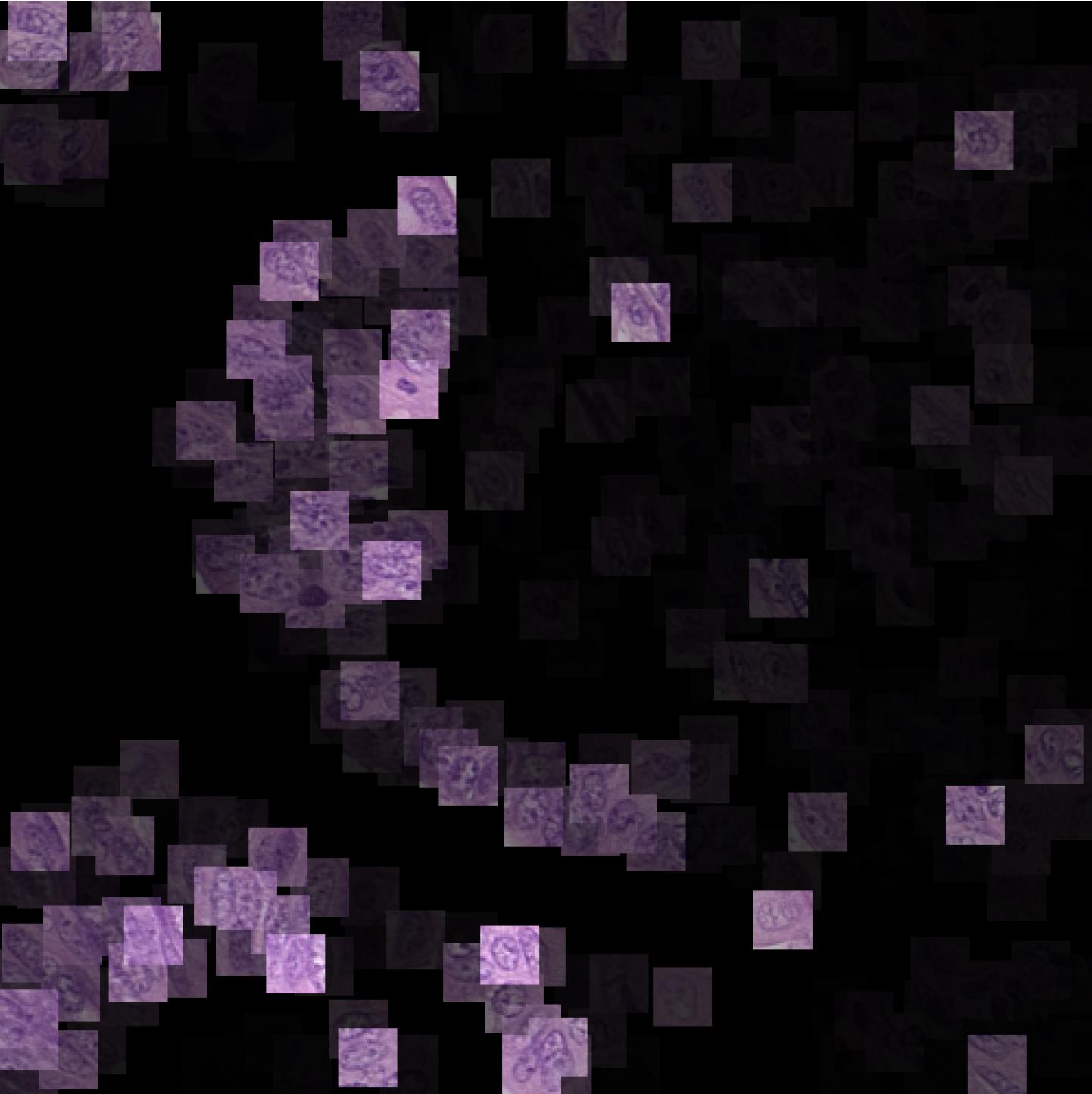}
		\caption{AttentionMIL}
	\end{subfigure}
	\begin{subfigure}{0.18\textwidth}
		\includegraphics[height=1.2in]{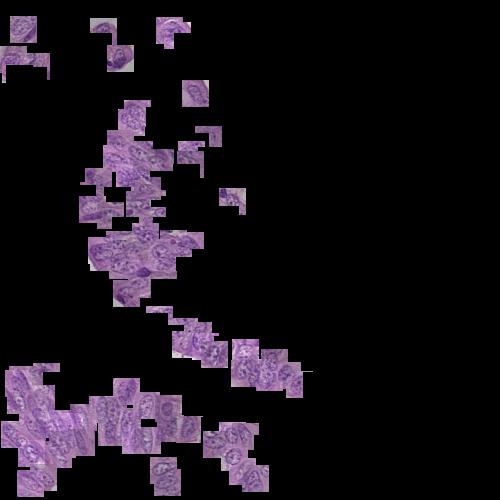}
		\caption{MIVAE}
	\end{subfigure}
	\caption{(a) Original H\&E stained images of the Colon Cancer dataset. (b) Ground truths. (c) Prediction heatmaps of mi-Net. (d) Prediction heatmaps of AttentionMIL. (e) Prediction heatmaps of MIVAE. This figure is best viewed in color.}
	\label{colon_cancer}
\end{figure*}
\subsection{Histopathology Images: Colon Cancer}
Computer-aided histopathology using hematoxylin and eosin (H\&E) stained whole-slide images is an important task for MIL since traditional supervised approaches require pixel-level annotations, which is extremely labor-intensive for pathologists \cite{Skrede2020}. 
Therefore, a label efficient multi-instance approach using only image-level labels is essential. 
For this purpose, we evaluate MIVAE against state-of-the-art end-to-end neural network-based MIL algorithms on both bag and instance label prediction using H\&E histopathology images from the colon cancer dataset provided in \cite{Sirinukunwattana2016}.

The dataset contains 100 H\&E images where each image is from a patient's tissue of either normal or malignant regions. 
For each image bag, the instances are generated as patches $27\times 27$ pixels using markings of major nuclei for each cell. A total amount of 22,444 nuclei instances are provided with ground truth instance class labels, i.e. epithelial, inflammatory, fibroblast, and miscellaneous.
The bag is considered to be positive if it contains at least one epithelial patch.

For fairness of comparison, the training images are color normalized and augmented with the same random horizontal flipping/vertical flipping, and random rotation procedures.
Furthermore, they also use the same convolutional structures for feature extraction (please refer to the supplementary for details).
When evaluating bag label prediction, we use accuracy since the bag labels are evenly distributed.
Because the instances label are imbalanced, we use AUC-PR for quantitative evaluating instance label prediction performance as in the last section. The reported results are the average performances from 5 times repeated 10-fold cross-validation.

As shown in the results of Table \ref{histopathology}, MIVAE achieves both the highest classification accuracy for predicting bag labels and the best AUC-PR scores for predicting instance labels, indicating that MIVAE outperforms state-of-the-art neural network-based MIL algorithms at both tasks. 

In Figure \ref{colon_cancer}, we use heatmaps to visually compare the instance label prediction performances among MIVAE against mi-Net and AttentionMIL. 
For each image, the heatmaps are obtained by multiplying the instances prediction scores of each algorithm (normalized to the range of $[0,1]$) to the pixel values of the image patches. 
From Figure \ref{colon_cancer} (b)-(e), it is evident that the heatmaps generated by MIVAE are the closest to the ground truths, while mi-Net exhibits lower recalls and the predictions of AttentionMIL are not as confident as the predictions of MIVAE.

It is worth noting that both mi-Net and AttentionMIL assume the instances to be independently distributed.
Moreover, mi-Net also uses max-pooling when linking instance prediction scores to bag label. 
Therefore, contrasting the heatmaps of MIVAE and mi-Net paints a clear picture for the benefits of modelling the instance dependencies: 
mi-Net only selects a small portion of positive patches, 
while MIVAE is able to identify most of them because the shared dependencies are excluded from instance label predictions.
Furthermore, comparing MIVAE with AttentionMIL shows that by using the instance-level factors, max-pooling provides more confident predictions than the attention weights.
For more examples, please kindly refer to the supplementary materials.

To sum up, the experiments results demonstrates that the proposed MIVAE algorithm is effective at both bag label and instance label prediction tasks. 
Moreover, the instance prediction scores of MIVAE can be used for explaining the region of interests
which is valuable for weakly-labeled tasks where fine-grained labels are costly to obtain.

\section{Conclusion}
In this paper, we propose MIVAE, a generative multi-instance learning algorithm which models the instances within a bag as non-i.i.d.. 
Experiments on several MIL benchmarks and medical imaging datasets demonstrate that MIVAE is effective for both instance-level and bag-level label prediction. 
An interesting future direction would be to investigate how other MIL pooling techniques affect the performance of MIVAE. Furthermore, adapting MIVAE to address the distribution change problem is also worth exploring.
\bibliographystyle{named}
\bibliography{reference}

\begin{thebibliography}{}

\bibitem[\protect\citeauthoryear{Amores}{2013}]{Amores2013}
Jaume Amores.
\newblock Multiple instance classification: Review, taxonomy and comparative
  study.
\newblock {\em Artificial Intelligence}, 201:81--105, 2013.

\bibitem[\protect\citeauthoryear{Andrews \bgroup \em et al.\egroup
  }{2002}]{Andrews:2002:SVM:2968618.2968690}
Stuart Andrews, Ioannis Tsochantaridis, and Thomas Hofmann.
\newblock Support vector machines for multiple-instance learning.
\newblock In {\em Proceedings of the 15th International Conference on Neural
  Information Processing Systems (NeurIPS '02)}, pages 577--584, 2002.

\bibitem[\protect\citeauthoryear{Chen \bgroup \em et al.\egroup
  }{2006}]{Chen2006}
Yixin Chen, Jinbo Bi, and J.Z. Wang.
\newblock {MILES}: Multiple-instance learning via embedded instance selection.
\newblock {\em {IEEE} Transactions on Pattern Analysis and Machine
  Intelligence}, 28(12):1931--1947, 2006.

\bibitem[\protect\citeauthoryear{Dietterich \bgroup \em et al.\egroup
  }{1997}]{Dietterich1997}
Thomas~G. Dietterich, Richard~H. Lathrop, and Tom{\'{a}}s Lozano-P{\'{e}}rez.
\newblock Solving the multiple instance problem with axis-parallel rectangles.
\newblock {\em Artificial Intelligence}, 89(1-2):31--71, 1997.

\bibitem[\protect\citeauthoryear{Doran and Ray}{2016}]{Doran2016}
Gary Doran and Soumya Ray.
\newblock Multiple-instance learning from distributions.
\newblock {\em Journal of Machine Learning Research}, 17(128):1--50, 2016.

\bibitem[\protect\citeauthoryear{Gärtner \bgroup \em et al.\egroup
  }{2002}]{Gaertner2002}
Thomas Gärtner, Peter A.Flach, Adam Kowalczyk, and Alex J.Smola.
\newblock Multi-instance kernels.
\newblock In {\em Proceedings of the 19th International Conference on Machine
  Learning (ICML '02)}, pages 179--186, 2002.

\bibitem[\protect\citeauthoryear{Haussmann \bgroup \em et al.\egroup
  }{2017}]{Haussmann2017}
Manuel Haussmann, Fred~A. Hamprecht, and Melih Kandemir.
\newblock Variational bayesian multiple instance learning with gaussian
  processes.
\newblock In {\em 2017 {IEEE} Conference on Computer Vision and Pattern
  Recognition ({CVPR '17})}. {IEEE}, 2017.

\bibitem[\protect\citeauthoryear{Ilse \bgroup \em et al.\egroup
  }{2018}]{Ilse2018}
Maximilian Ilse, Jakub Tomczak, and Max Welling.
\newblock Attention-based deep multiple instance learning.
\newblock In {\em Proceedings of the 35th International Conference on Machine
  Learning (ICML '18)}, pages 2127--2136, 2018.

\bibitem[\protect\citeauthoryear{Ji \bgroup \em et al.\egroup }{2020}]{Ji2020}
Yunjie Ji, Hao Liu, Bolei He, Xinyan Xiao, Hua Wu, and Yanhua Yu.
\newblock Diversified multiple instance learning for document-level
  multi-aspect sentiment classification.
\newblock In {\em Proceedings of the 2020 Conference on Empirical Methods in
  Natural Language Processing (EMNLP '20)}, pages 7012--7023, 2020.

\bibitem[\protect\citeauthoryear{Kandemir and Hamprecht}{2014}]{Kandemir2014}
Melih Kandemir and Fred~A. Hamprecht.
\newblock Instance label prediction by dirichlet process multiple instance
  learning.
\newblock In {\em Proceedings of the Thirtieth Conference on Uncertainty in
  Artificial Intelligence (UAI '14)}, 2014.

\bibitem[\protect\citeauthoryear{Kim and la Torre}{2010}]{Kim2010}
Minyoung Kim and Fernando~De la~Torre.
\newblock Gaussian processes multiple instance learning minyoung.
\newblock In {\em Proceedings of the 27th International Conference on Machine
  Learning (ICML '10)}, 2010.

\bibitem[\protect\citeauthoryear{Kingma and Welling}{2014}]{Kingma2014}
Diederik~P Kingma and Max Welling.
\newblock Auto-encoding variational bayes.
\newblock In {\em Proceedings of the 2nd International Conference on Learning
  Representations (ICLR '14)}, 2014.

\bibitem[\protect\citeauthoryear{Li \bgroup \em et al.\egroup }{2009}]{Li2009}
Yu-Feng Li, James T.Kwok, Ivor~W. Tsang, and Zhi-Hua Zhou.
\newblock A convex method for locating regions of interest with multi-instance
  learning.
\newblock In {\em Proceedings of the European Conference on Machine Learning
  and Principles and Practice of Knowledge Discovery in Databases (ECML PKDD
  '09)}, 2009.

\bibitem[\protect\citeauthoryear{Shi \bgroup \em et al.\egroup
  }{2020}]{Shi2020}
Xiaoshuang Shi, Fuyong Xing, Yuanpu Xie, Zizhao Zhang, Lei Cui, and Lin Yang.
\newblock Loss-based attention for deep multiple instance learning.
\newblock {\em Proceedings of the 34th {AAAI} Conference on Artificial
  Intelligence (AAAI '20)}, pages 5742--5749, 2020.

\bibitem[\protect\citeauthoryear{Sirinukunwattana \bgroup \em et al.\egroup
  }{2016}]{Sirinukunwattana2016}
Korsuk Sirinukunwattana, Shan E~Ahmed Raza, Yee-Wah Tsang, David R.~J. Snead,
  Ian~A. Cree, and Nasir~M. Rajpoot.
\newblock Locality sensitive deep learning for detection and classification of
  nuclei in routine colon cancer histology images.
\newblock {\em {IEEE} Transactions on Medical Imaging}, 35(5):1196--1206, may
  2016.

\bibitem[\protect\citeauthoryear{Skrede \bgroup \em et al.\egroup
  }{2020}]{Skrede2020}
Ole-Johan Skrede, Sepp~De Raedt, Andreas~Kleppe andTarjei S~Hveem, Knut
  Liestøl, John Maddison, Hanne~A Askautrud, Manohar Pradhan, John~Arne
  Nesheim, Fritz Albregtsen, Inger~Nina Farstad, Enric Domingo, David~N Church,
  Arild Nesbakken, Neil~A Shepherd, Ian Tomlinson, Rachel Kerr, Marco Novelli,
  David~J Kerr, and Håvard~E Danielsen.
\newblock Deep learning for prediction of colorectal cancer outcome: a
  discovery and validation study.
\newblock {\em The Lancet}, 35(10221):350--360, 2020.

\bibitem[\protect\citeauthoryear{Wang \bgroup \em et al.\egroup
  }{2018}]{Wang2018}
Xinggang Wang, Yongluan Yan, Peng Tang, Xiang Bai, and Wenyu Liu.
\newblock Revisiting multiple instance neural networks.
\newblock {\em Pattern Recognition}, 74:15--24, 2018.

\bibitem[\protect\citeauthoryear{Wei \bgroup \em et al.\egroup
  }{2017}]{Wei2017}
Xiu-Shen Wei, Jianxin Wu, and Zhi-Hua Zhou.
\newblock Scalable algorithms for multi-instance learning.
\newblock {\em {IEEE} Transactions on Neural Networks and Learning Systems},
  28(4):975--987, 2017.

\bibitem[\protect\citeauthoryear{Xu \bgroup \em et al.\egroup }{2019}]{Xu2019}
Bi-Cun Xu, Kai~Ming Ting, and Zhi-Hua Zhou.
\newblock Isolation set-kernel and its application to multi-instance learning.
\newblock In {\em Proceedings of The 25th ACM SIGKDD Conference on Knowledge
  Discovery and Data Mining (KDD '19)}, pages 941--949, 2019.

\bibitem[\protect\citeauthoryear{Yao \bgroup \em et al.\egroup
  }{2020}]{Yao2020}
Jiawen Yao, Xinliang Zhu, Jitendra Jonnagaddala, Nicholas Hawkins, and Junzhou
  Huang.
\newblock Whole slide images based cancer survival prediction using attention
  guided deep multiple instance learning networks.
\newblock {\em Medical Image Analysis}, 65:101789, oct 2020.

\bibitem[\protect\citeauthoryear{Zhang and Zhou}{2014}]{Zhang2014}
Weijia Zhang and Zhi-Hua Zhou.
\newblock Multi-instance learning with distribution change.
\newblock In {\em Proceedings of the 28th AAAI Conference on Artificial
  Intelligence (AAAI '14)}, pages 2184--2190, 2014.

\bibitem[\protect\citeauthoryear{Zhou \bgroup \em et al.\egroup
  }{2009}]{Zhou2009}
Zhi-Hua Zhou, Yu-Yin Sun, and Yu-Feng Li.
\newblock Multi-instance learning by treating instances as non-i.i.d. samples.
\newblock In {\em Proceedings of the 26th International Conference on Machine
  Learning (ICML '09)}, pages 1249--1256, 2009.

\end{thebibliography}

\onecolumn
\section{Network Structures and Parameters}
The experiments are almost entirely performed using a personal computer equipped with a AMD Ryzen\texttrademark \space 3700X CPU, 32GB of memory, and a Nvidia Geforce\textregistered \space 1080Ti GPU with 11GB of memory. 
The code of MIVAE is implemented using PyTorch 1.6. For all experiments, the models are training with AdamW for 100 epochs. The parameters and best epochs are selected using the loss and accuracy on the validation data, which is consisted of 10\% of bags reserved from the training data. 

\subsection{Benchmark Multi-instance datasets}
For the benchmark datasets including Musk1, Musk2, Fox, Elephant, Tiger and all of the 20 NewsGroup datasets, we use the same network structure. Specifically, all the encoders and decoders are full-connected neural networks with the same number of hidden layers and the same number of hidden units. Between each layer, we use the ReLU function for activation. We list the grid of searched parameters in Table \ref{parameters_benchmark}. 
\begin{table}[!h]
	\centering
	\caption{Parameter grids for benchmark datasets}
	\begin{tabular}{|l|l|}
		\hline
		Number of hidden layers & [2,3] \\
		\hline
		Number of hidden units & [100,200]\\
		\hline
		Latent dim $D_{z_B}$, $D_{z_I}$ & [16,32,64] \\
		\hline
		Auxiliary weight $\alpha$ & [100, 1000, 10000]\\
		\hline
		Learning rate & [1e-3, 1e-4] \\
		\hline
		Weight decay & [1e-2, 1e-3, 1e-4] \\
		\hline
	\end{tabular}
	\label{parameters_benchmark}
\end{table}

\subsection{Colon Cancer}
For the Colon Cancer histopathology dataset, we use the same convoluntional network structure as the compared methods for fairness of comparison. For encoding the images, we used the following convolutional network (Table \ref{conv} (left)) as used in \cite{Ilse2018}. For decoding the from the latent factors, we use a corresponding de-convolutional network as specified in Table \ref{conv} (right). All other encoders and decoders are full-connected neural networks as used in the benchmark datasets. The parameters grids for image datasets are provided in Table \ref{parameters_image}.

\begin{table}[h]
	\centering
	\caption{Convolution and de-convolution networks.}
	\begin{tabular}{|c|c|}
		\hline
		Layer & Type \\
		\hline
		1 & conv(4,1,0)-36 + ReLU\\
		\hline
		2 & maxpool(2,2) \\
		\hline
		3 & conv(3,1,0)-48 + ReLU\\
		\hline
		4 & maxpool(2,2) \\
		\hline
	\end{tabular}
	~
	\begin{tabular}{|c|c|}
		\hline
		Layer & Type \\
		\hline
		1 & Upsampling(10)\\
		\hline
		2 & ConvTranspose(3,1,0)-48 +ReLU\\
		\hline
		3 & Upsampling(5)\\
		\hline
		4 &   ConvTranspose(4,1,0)-36 +ReLU \\
		\hline
	\end{tabular}
	\label{conv}
\end{table}

\begin{table}[!h]
	\centering
	\caption{Parameter grids for Colon Cancer dataset.}
	\begin{tabular}{|l|l|}
		\hline
		Number of hidden layers & [3,4,5] \\
		\hline
		Number of hidden units & [256,512]]\\
		\hline
		Latent dim $D_{z_B}$, $D_{z_I}$ & [32,64,128] \\
		\hline
		Auxiliary weight $\alpha$ & [100, 1000, 10000]\\
		\hline
		Learning rate & [1e-3, 5e-4, 1e-4] \\
		\hline
		Weight decay & [1e-3, 5e-4, 1e-4] \\
		\hline
	\end{tabular}
	\label{parameters_image}
\end{table}

\section{Data Availability}
All of the datasets used in the paper is publicly available online. The Musk1, Musk2, Fox, Elephant, and Tiger datasets can be accessed at \url{http://homepage.tudelft.nl/n9d04/milweb/index.html}.
The 20 NewsGroups datasets can be downloaded from \url{http://www.lamda.nju.edu.cn/data_MItext.ashx}. 

The original images and labels of the Colon Cancer dataset can be accessed at \url{https://warwick.ac.uk/fac/sci/dcs/research/tia/data/crchistolabelednucleihe/}. Furthermore, the processed multi-instance bags can be downloaded directly from \url{https://drive.google.com/file/d/1RcNlwg0TwaZoaFO0uMXHFtAo_DCVPE6z/view?usp=sharing}. 

The original images for the Pediatric Chest X-Ray dataset are available from \url{https://www.kaggle.com/paultimothymooney/chest-xray-pneumonia}. The script for generating the patches are available at \url{https://github.com/JonGerrand/Deep-Multi-Instance-Learning}.

\section{More examples from Colon Cancer dataset}
\begin{figure*}[!h]
	\centering
	\begin{subfigure}{0.18\textwidth}
		\includegraphics[height=1in]{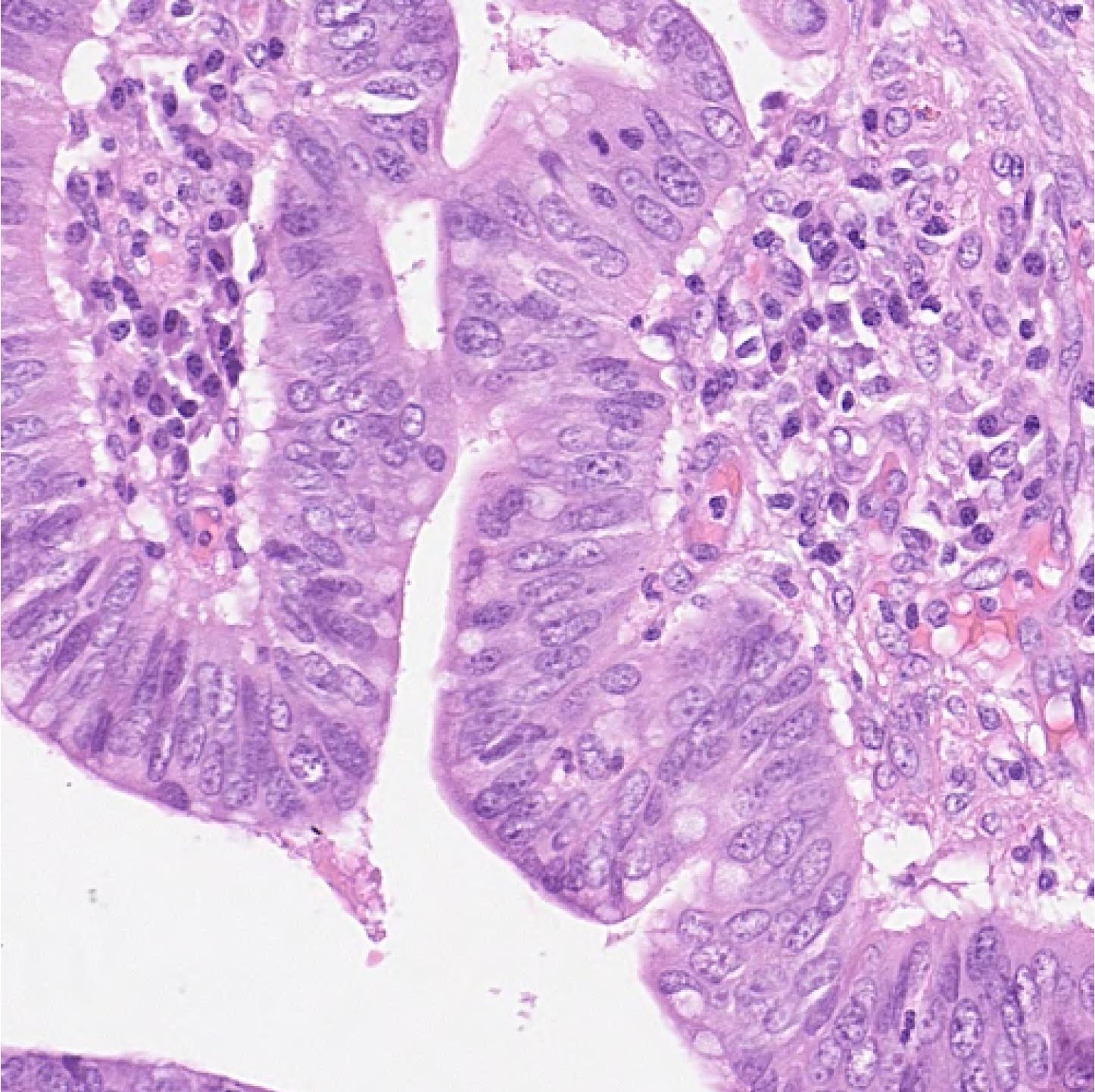}
		\caption{Original image}
	\end{subfigure}
	\begin{subfigure}{0.18\textwidth}
		\includegraphics[height=1in]{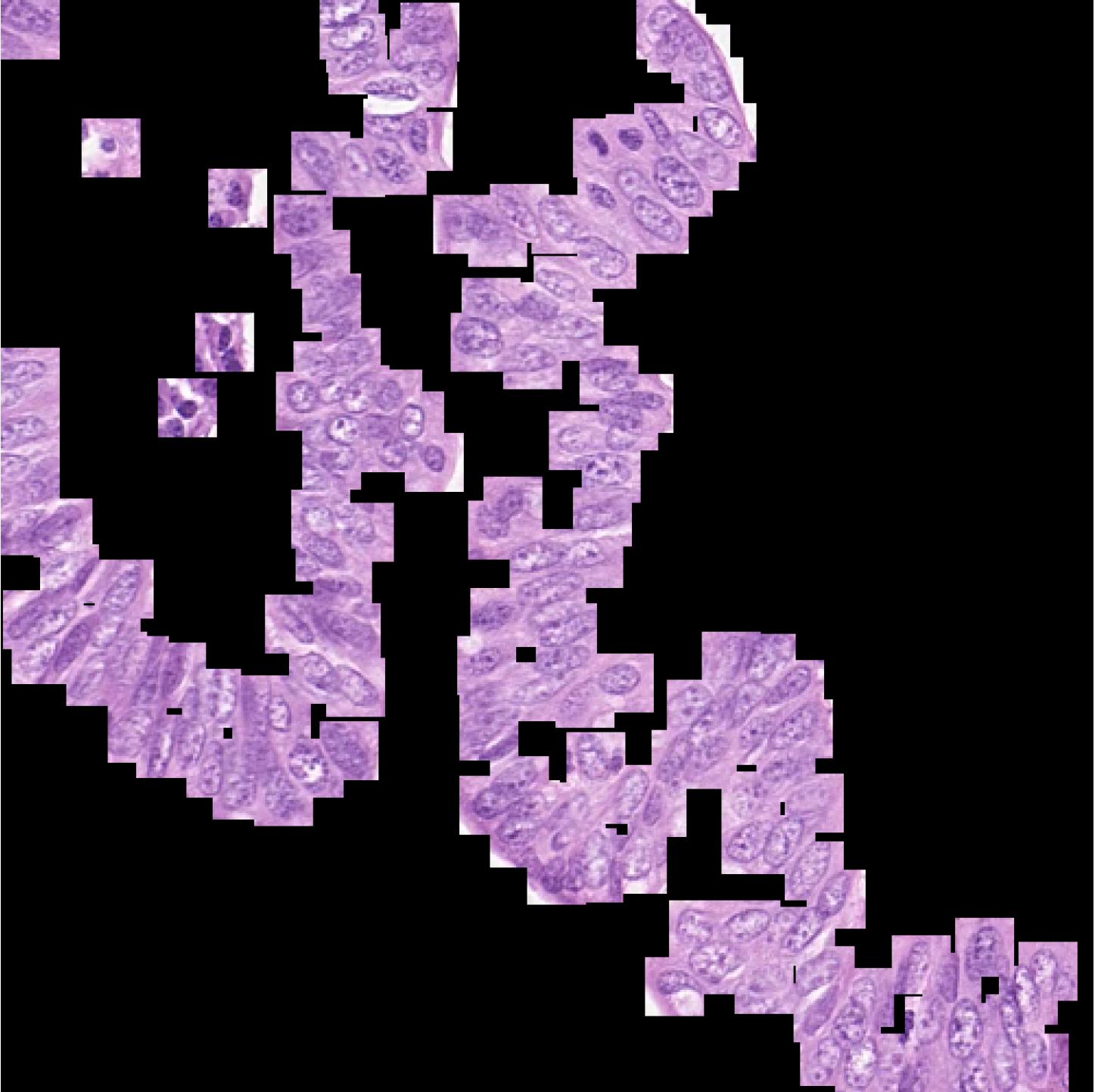}
		\caption{Ground truth}
	\end{subfigure}
	\begin{subfigure}{0.18\textwidth}
		\includegraphics[height=1in]{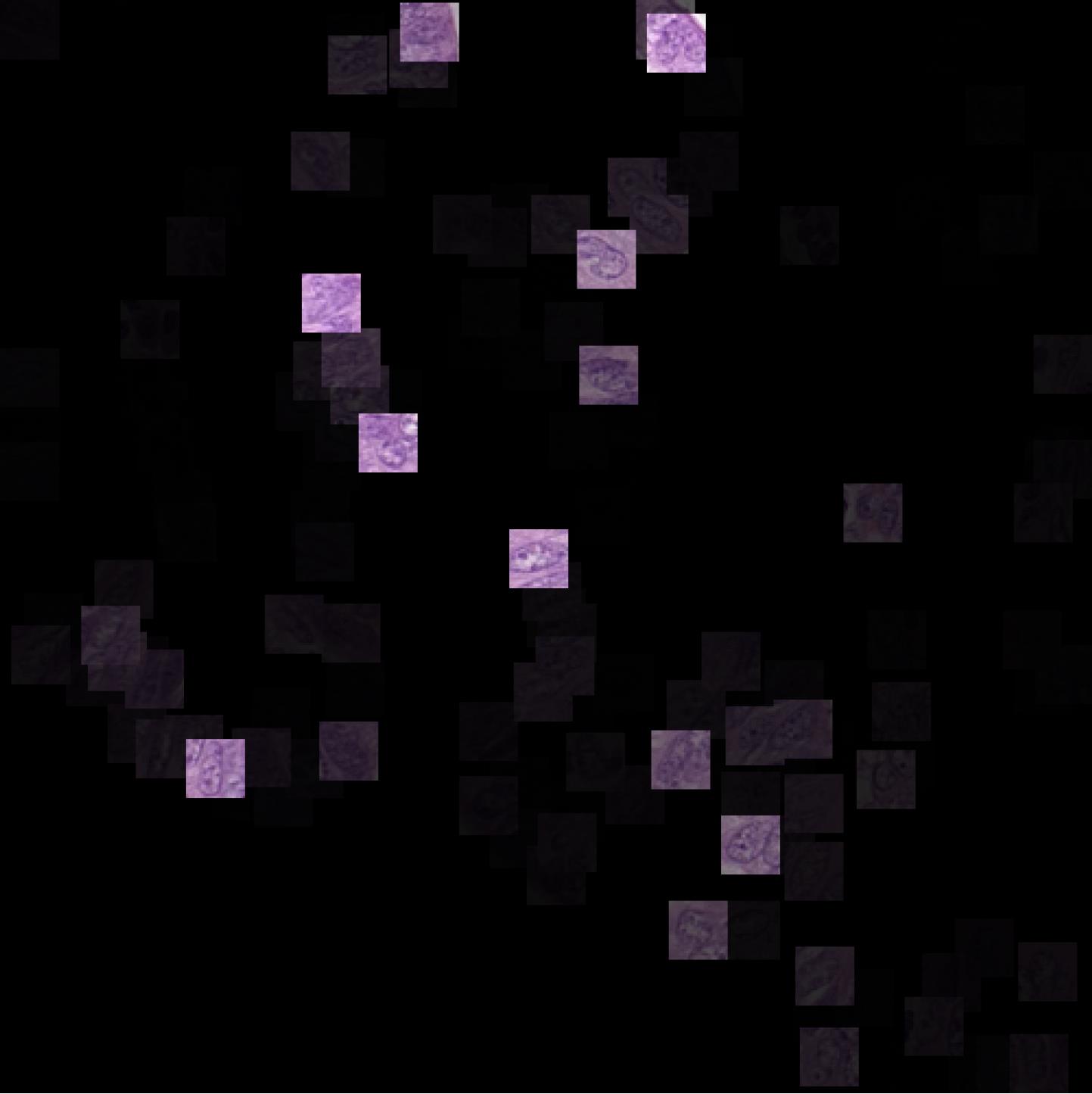}
		\caption{mi-Net}
	\end{subfigure}
	\begin{subfigure}{0.18\textwidth}
		\includegraphics[height=1in]{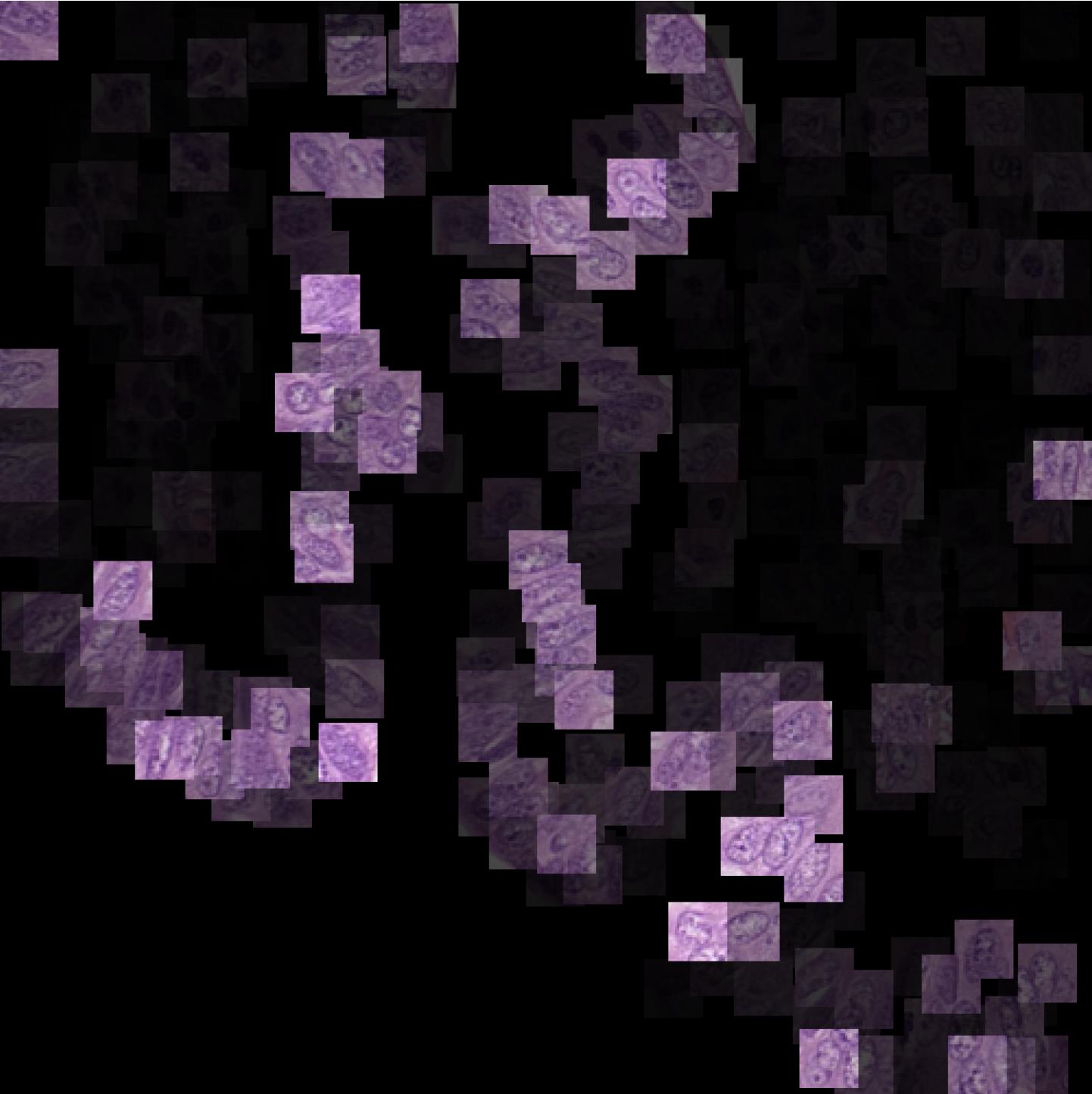}
		\caption{AttentionMIL}
	\end{subfigure}
	\begin{subfigure}{0.18\textwidth}
		\includegraphics[height=1in]{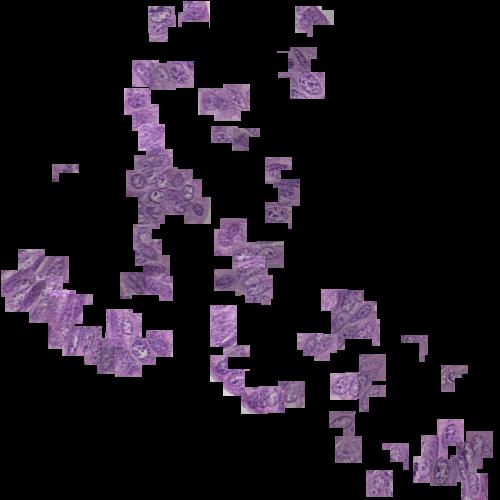}
		\caption{MIVAE}
	\end{subfigure}
	\caption{Colon Cancer example 1: (a) Original H\&E stained images from the Colon Cancer dataset. (b) Ground truth instances labels. (c) Predicting heatmap of mi-Net (max pooling). (d) Prediction heatmap of AttentionMIL. (e) Prediction heatmap of MIVAE (max pooling). }
	\label{colon_cancer}
\end{figure*}

\begin{figure*}[!h]
	\centering
	\begin{subfigure}{0.18\textwidth}
		\includegraphics[height=1in]{img10_original.jpg}
		\caption{Original image}
	\end{subfigure}
	\begin{subfigure}{0.18\textwidth}
		\includegraphics[height=1in]{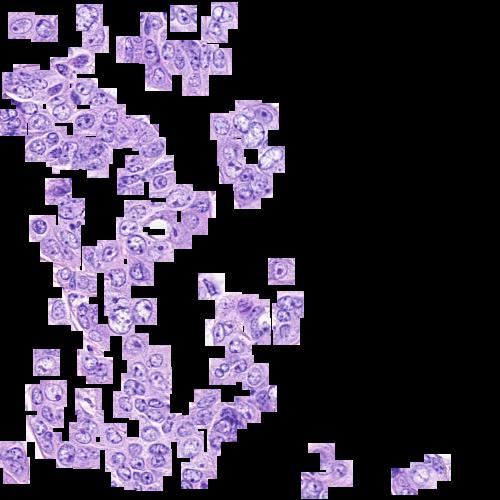}
		\caption{Ground truth}
	\end{subfigure}
	\begin{subfigure}{0.18\textwidth}
		\includegraphics[height=1in]{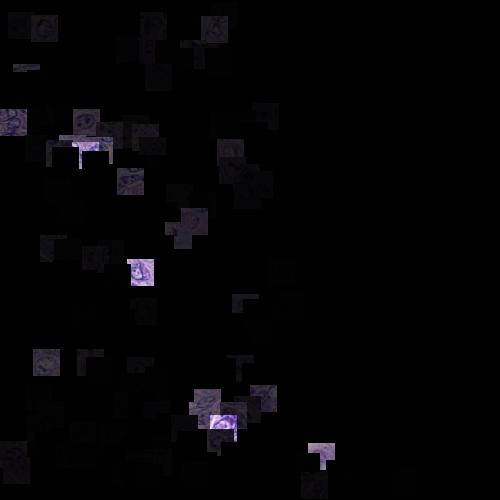}
		\caption{mi-Net}
	\end{subfigure}
	\begin{subfigure}{0.18\textwidth}
		\includegraphics[height=1in]{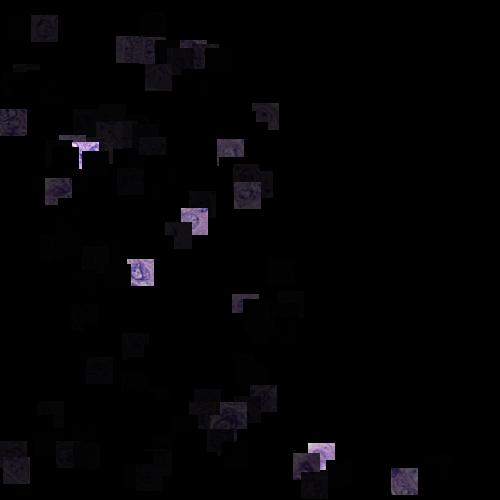}
		\caption{AttentionMIL}
	\end{subfigure}
	\begin{subfigure}{0.18\textwidth}
		\includegraphics[height=1in]{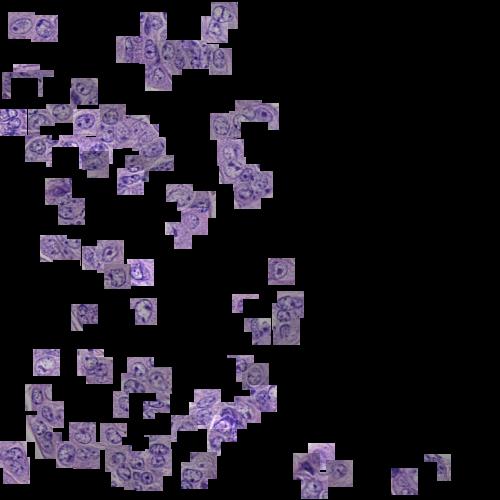}
		\caption{MIVAE}
	\end{subfigure}
	\caption{Colon Cancer example 2: (a) Original H\&E stained images from the Colon Cancer dataset. (b) Ground truth instances labels. (c) Predicting heatmap of mi-Net (max pooling). (d) Prediction heatmap of AttentionMIL. (e) Prediction heatmap of MIVAE (max pooling). }
	\label{colon_cancer2}
\end{figure*}
\end{document}